\RequirePackage{fix-cm}
\documentclass[twocolumn]{svjour3}          % twocolumn
\smartqed  % flush right qed marks, e.g. at end of proof
\usepackage{graphicx}
\usepackage{times}
\usepackage{epsfig}
\usepackage{epstopdf}
\usepackage{amsmath}
\usepackage{amssymb}
\usepackage{subfigure}
\usepackage{caption}
\usepackage{tabularx}
\usepackage[misc]{ifsym}

\usepackage{booktabs}
\usepackage{multirow}
\usepackage{makecell}
\usepackage{array}
\usepackage{algorithmic}
\usepackage{tabularx}

\usepackage[table]{xcolor}
 \usepackage[normalem]{ulem}
\usepackage{algorithm}

\usepackage[section]{placeins}
\usepackage{natbib}
\setcitestyle{authoryear,round}
\makeatletter
\def\BState{\State\hskip-\ALG@thistlm}
\makeatother
\definecolor{myGreen}{HTML}{33FF00}
\definecolor{myRed}{HTML}{FF3030}
\definecolor{myGrey}{HTML}{AA5555}
\definecolor{myWhite}{HTML}{FFFFFF}
\definecolor{maroon}{cmyk}{0,0.87,0.68,0.32}
\definecolor{petr}{HTML}{5555FF}
\definecolor{josef}{HTML}{FF3030}
\usepackage{ragged2e}

\renewcommand{\algorithmicrequire}{\textbf{Input:}}
\renewcommand{\algorithmicensure}{\textbf{Output:}}
\journalname{IJCV}
\begin{document}
\title{Relation-Guided Adversarial Learning for Data-free Knowledge Transfer}

%\titlerunning{Short form of title}        % if too long for running head

\author{Yingping Liang         \and
        Ying Fu %etc.
}

%\authorrunning{Short form of author list} % if too long for running head

\institute{Yingping Liang \at
              School of Computer Science and Technology, Beijing Institute of Technology, Beijing, China. \\
              \email{liangyingping@bit.edu.cn}           %  \\
%             \emph{Present address:} of F. Author  %  if needed
           \and
           \Letter  Ying Fu \at
              School of Computer Science and Technology, Beijing Institute of Technology, Beijing, China. \\
              \email{fuying@bit.edu.cn}  
}
\vspace{-5mm}
\date{Received: date / Accepted: date}

\maketitle

\begin{abstract}
    Data-free knowledge distillation transfers knowledge by recovering training data from a pre-trained model. Despite the recent success of seeking global data diversity, the diversity within each class and the similarity among different classes are largely overlooked, resulting in data homogeneity and limited performance. In this paper, we introduce a novel \textbf{R}elation-\textbf{G}uided \textbf{A}dversarial \textbf{L}earning method with triplet losses, which solves the homogeneity problem from two aspects. To be specific, our method aims to promote both intra-class diversity and inter-class confusion of the generated samples. To this end, we design two phases, an image synthesis phase and a student training phase. In the image synthesis phase, we construct an optimization process to push away samples with the same labels and pull close samples with different labels, leading to intra-class diversity and inter-class confusion, respectively. Then, in the student training phase, we perform an opposite optimization, which adversarially attempts to reduce the distance of samples of the same classes and enlarge the distance of samples of different classes. To mitigate the conflict of seeking high global diversity and keeping inter-class confusing, we propose a focal weighted sampling strategy by selecting the negative in the triplets unevenly within a finite range of distance. RGAL shows significant improvement over previous state-of-the-art methods in accuracy and data efficiency. Besides, RGAL can be inserted into state-of-the-art methods on various data-free knowledge transfer applications. Experiments on various benchmarks demonstrate the effectiveness and generalizability of our proposed method on various tasks, specially data-free knowledge distillation, data-free quantization, and non-exemplar incremental learning. Our code will be publicly available to the community.
\end{abstract}

\section{Introduction}
\label{intro}
Deep convolutional neural networks (CNNs) have demonstrated remarkable success in various computer vision tasks, such as classification \citep{krizhevsky2012imagenet, simonyan2014very, he2016deep}, object detection \citep{girshick2014rich, redmon2016you, zhang2018single}, and semantic segmentation \citep{long2015fully, ronneberger2015u}. However, these well-designed models require significant computational resources, which limits their deployment on resource-limited platforms such as mobile phones and edge devices \citep{cheng2018model}.

To address this issue, knowledge distillation (KD) has emerged as a standard model compression method that aims to transfer knowledge from a well-trained teacher model to a new, smaller one \citep{gou2021knowledge, hinton2015distilling, romero2014fitnets}. Despite its success, most KD methods assume that the training dataset is still accessible. However, in many cases, data privacy and security concerns prevent the sharing of sensitive data \citep{ha2020security}. For instance, biometric and medical data are often considered private and not publicly shared \citep{arora2020biometric}. Therefore, there is a pressing need to transfer knowledge without accessing the original dataset or the real data.

Data-free KD can effectively solve the challenge of transferring knowledge between neural networks without access to the original training data. Inspired by the generative adversarial network \citep{goodfellow2014generative}, existing data-free KD methods typically generate synthetic samples that the teacher classifies with high certainty \citep{mahendran2015understanding}. Subsequently, adversarial learning is introduced into the field of data-free knowledge distillation to explore more valuable samples by minimizing the maximum KL divergence between the outputs of the teacher and the student \citep{Micaelli2019ZeroShotKT, choi2020data}. However, real-world images of different classes often contain different contents, while synthesized samples tend to be homogenized, leading to limited performance of models trained on these synthesized samples. It is essential for the synthesized samples to sufficiently match the diversity in real data.

Therefore, recent data-free methods try to enhance the diversity of synthetic data. For example, Deepinversion \citep{yin2020dreaming} introduces an additional Jensen-Shannon divergence to expand the coverage of the image distribution. IntraQ \citep{zhong2021intraq} proposes diversity-seeking regularization to prevent the generated samples from being too similar. CMI \citep{fang2021contrastive} improves diversity with a large amount of pre-synthesized images and encourages the synthesizing images to distinguish from the previous ones. 

Although these methods enhance the diversity of synthetic data, models trained with real data still outperform those trained with synthetic data. This may be because current methods often overlook the relation among individual samples and optimize the samples independently in the same direction. This causes synthetic samples to tend to have similar results without avoiding homogenization problems. Intra-class samples cluster together, and inter-class samples become too easily distinguishable. Therefore, we emphasize that it is necessary to improve the diversity of individual samples within the same training batch, especially those of the same classes. In addition, our study has also revealed class-level heterogeneity issues in the data generation process of typical data-free methods, particularly at embedding levels. Therefore, the lack of hard samples with high inter-class confusion also prevents the student from learning more robust class-level discrimination. These gaps lead to difficulties in learning from synthetic data with high homogeneity.

\begin{figure}[t]
   \centering
   \includegraphics[width=0.48\textwidth]{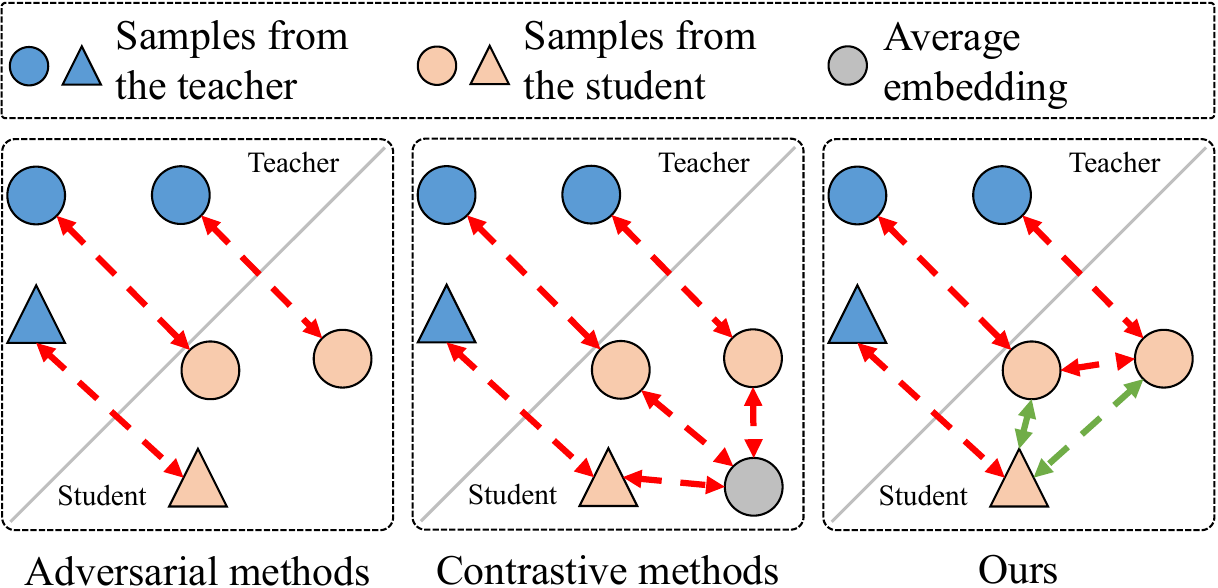} 
   \caption{Illustration of sample optimization in different data generation methods. Red arrow indicates pushing away while green indicates pulling close. Both adversarial based and contrastive methods ignore the relation between individual samples, resulting in limited intra-class diversity and inter-class confusion. Our method aims to deal with the distances among samples, leading to both high intra-class diversity and meanwhile maintaining inter-class confusion.}
   \label{fig:teaser}
\end{figure}

To address these problems, we propose a novel Relation-Guided Adversarial Learning (RGAL) framework by promoting both intra-class diversity and inter-class confusion. Unlike prior methods that optimize the samples independently, we focus on optimizing individual samples and the relation among the samples in a single batch.  Our proposed method refers to two learning phases, an image synthesis phase for training the generator and a student training phase. In the image synthesis phase, we enlarge the distance between positive pairs of the same classes in the embedding space, which prevents the samples from being too similar and ensures intra-class diversity. Negative pairs of different classes are forced to be close to each other, which ensures that the samples of different classes are around the decision boundary and "confusing" enough. On the contrary, when training the student, the positive pairs are pulled closer than the negative ones by a distance margin to learn much stronger class-level discrimination.

Figure \ref{fig:teaser} illustrates the comparison between the knowledge transfer processes using different data-free knowledge distillation (KD) methods: adversarial methods \cite{Micaelli2019ZeroShotKT, choi2020data, yin2020dreaming}, contrastive methods \cite{zhong2021intraq, fang2021contrastive, fang2022up}, and our proposed method. Adversarial methods generate synthetic samples classified with high uncertainty by adversarially learning pseudo data with larger gaps between the teacher and the student. However, the samples tend to be homogenized. Contrastive methods enhance diversity by driving the samples away from the previous ones. But the optimization of relationships among samples within a batch is also overlooked. In addition, as the number of samples increases, the average embedding tends to the class center, also resulting in homogenization problem. In contrast, our method focuses on relationships between individual samples and focuses on two different aspects: intra-class diversity and inter-class confusion. Therefore, we design triplet adversarial losses, which performs different optimizations between individual samples with different class labels in the image synthesis phase.

In the image synthesis phase, it also harms the diversity of global data to pull closer negative samples that have already been far apart or close enough. To maintain high global diversity, we propose a focal weighted sampling strategy to get the applicable negative for the triplet. To be specific, we sample the negative unevenly in terms of the inverse distance only in local views. Only those neither too far nor too close are likely to be sampled as the negative. This mitigates the conflicts of seeking global diversity and enhancing inter-class confusion. 

Our main contributions can be summarized as follows:
\begin{itemize}
\item We propose a relation-guided adversarial learning method to improve both intra-class diversity and inter-class confusion of individual samples within a batch and guide the student to learn robust class-level discrimination from generated samples without any real data.
\item We present a focal weighted strategy for unevenly sampling negative samples within local views, which mitigates the latent optimization conflicts between global diversity and inter-class class confusion.
\item Our method achieves more competitive accuracy than current methods on data-free knowledge distillation and significantly improves the performance of prior works on other tasks, such as data-free quantization and non-exemplar incremental learning.
\end{itemize}

\section{Related Work}

\noindent\textbf{Knowledge Transfer} is a traditional method to transfer the knowledge from a well-trained teacher model to a new student model \citep{bucilua2006model, ba2013deep} by smoothing the teacher’s probability outputs by temperature term as the learning guide for the student. Knowledge distillation \citep{hinton2015distilling} further matches the smooth distributions of the teacher and student models via Kullback-Leibler divergence. Fine-grained information among different layers also provides extra supervision to improve the student model performance, as mentioned by the work in \citep{romero2014fitnets}. However, these pre-trained models are sometimes released without training data for privacy or storage reasons, making these methods inapplicable. 

\noindent\textbf{Data-free Knowledge Distillation} attempts to address data limitation by re-constructing input samples from the parameter of a pre-trained teacher model \citep{lopes2017data}. ZSKD \citep{nayak2019zero} obtains the synthetic samples by directly optimizing trainable random noise images concerning a predetermined objective for multiple iterations. CMI \citep{fang2021contrastive} also re-visits the contrastive learning framework to model the data diversity in model inversion. Fast-DFKD \citep{fang2022up} inherits this idea and introduces a meta-learner to higher data synthesis efficiency. Deepinversion \citep{yin2020dreaming} improves the diversity of synthesized images by maximizing the Jensen-Shannon divergence of the outputs of the teacher and the student. Furthermore, RDSKD \citep{han2021robustness} generates images with high inter-sample diversity and carefully designs generator loss. However, all these methods ignore the contribution of hard samples that are inter-class confusing, which may be more critical to student learning. By extending these preceding works, we develop RGAL that focuses on separated intra-class diversity and inter-class confusion, which starkly contrasts prior works in this area (Section \ref{sec:dfkd}).

\noindent\textbf{Adversarial Learning} for image synthesis is introduced by GANs \citep{goodfellow2014generative}, where generators are used to produce adversarial images. Then, adversarial learning is introduced into data-free KD by DFAD \citep{fang2019data} and ZSKT \citep{Micaelli2019ZeroShotKT}, where worst-case samples with a large gap between the outputs of the student and the teacher are synthesized for student learning. The work in \citep{nie2020adversarial} proposes an adversarial confidence learning framework better to retain the unlimited modeling capacity of the discriminator. Nevertheless, traditional adversarial learning ignores the distance among individual samples. Thus the generator tends to generate samples clustering in the embedding space, which decreases the sample diversity within the same batch. To this end, we present a relation-guided adversarial learning method to diversify batch samples by the structural relation among samples with a focal weighted sampling strategy. 

\noindent\textbf{Data-Free Quantization} performs network quantization by extending the data-free knowledge distillation, where the models before and after quantification are regarded as the student and the teacher. For example, DFQAD \citep{choi2020data} minimizes the maximum distance between the outputs of the teacher and the (quantized) student from a generator. ZeroQ \citep{cai2020zeroq} matches the statistical distribution of the generated samples with the mean/standard deviation parameters stored in the BN layer. ZAQ \citep{liu2021zero} synthesizes informative and diverse samples with a two-level discrepancy modeling to drive a generator. Moreover, IntraQ \citep{zhong2021intraq} introduces a marginal distance constraint to form class-related embeddings distributed to retain this property in intra-class diversity in synthetic images. RGAL can be used as an easy-to-use trick to improve the accuracy of the quantized model by modifying the training losses of existing data-free quantization methods (Section \ref{sec:dfq}).

\noindent\textbf{Non-Exemplar Incremental Learning} tries to train the old model to a new one to recognize the old and new classes without the old class samples. In the paradigm of KD, the teacher, and the student is the old and new models, respectively. More formally, an incremental learning problem $T$ consists of a sequence of n tasks:
\begin{equation}
  \text{Tasks}=\left[\left(\mathrm{C}^{1}, \mathrm{D}^{1}\right),\left(\mathrm{C}^{2}, \mathrm{D}^{2}\right), \ldots,\left(\mathrm{C}^{\mathrm{n}}, \mathrm{D}^{\mathrm{n}}\right)\right].
\end{equation}
where each task $t$ is represented by a set of classes $\mathrm{C}^{\mathrm{t}}=\left\{\mathrm{c}_{1}^{\mathrm{t}},\mathrm{c}_{2}^{\mathrm{t}} \ldots,\mathrm{c}_{\mathrm{n}}^{\mathrm{t}}\right\}$ and training data $D^t$ in which:
\begin{equation}
 D^{t}=\left\{\left(x_{1}, y_{1}\right),\left(x_{2}, y_{2}\right), \ldots,\left(x_{m} t, y_{m} t\right)\right\}.
\end{equation}
The teacher $\mathcal{T}$ is initially trained on $[D^1, \dots, D^{t-1}]$ with classes $\mathcal{C}_o$. During training for task $t$, the learner (effectively as the student) only has access to $Dt$, and the tasks do not overlap in classes, i.e., $C^{i} \cap C^{j}=\varnothing \text { if } i \neq j$. PASS \citep{zhu2021prototype} proposes to memorize one class representative prototype and adopts prototype augmentation by gaussian noise to sample the representation of old classes. SSRE \citep{zhu2022self} enhances the discrimination between the old and new classes by selectively incorporating new samples. However, existing methods use the manual design of sample sampling, lacking the ability to adaptively generate the training samples required by the student (new) model. We extend the existing non-exemplar incremental learning methods based on RGAL, which can significantly improve the performance in terms of both the average accuracy and the average forgetting with just a few more optimization steps (Section \ref{sec:dfil}).

\section{Method}

\noindent\textbf{Motivation.} Prior works optimize the global class diversity of samples with class semantic embedding, neglecting the instance-level mutual information for knowledge transfer. Some works enlarge the distances between samples with the same labels to increase intra-class diversity, where the similarity among classes has been largely overlooked. To properly generate synthetic samples for efficient knowledge transfer, we propose Relation-Guided Adversarial Learning (RGAL). Our motivation is intuitive that under the constraint of data amount, higher intra-class diversity indicates better instance-level representation. Besides, higher inter-class confusion indicates that stronger class-level discrimination can be learned from the teacher by the student. 

\begin{figure*}[t]
   \centering
   \includegraphics[width=0.95\textwidth]{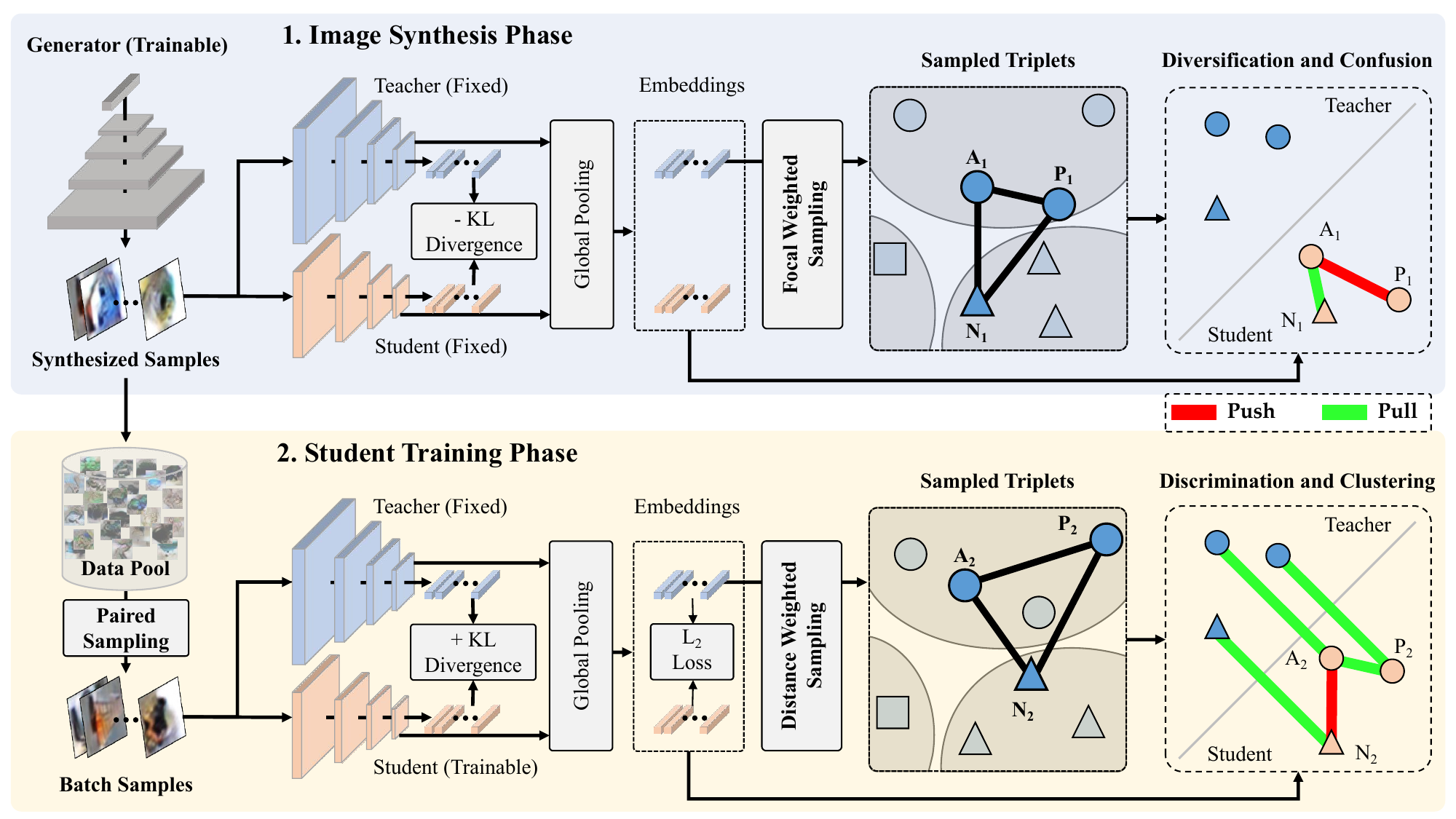} 
   \caption{The framework of our proposed RGAL method for data-free knowledge distillation. A, P, and N represent the anchor sample, positive sample, and negative sample, respectively. Our method alternates training the generator and the student model using triplet losses in opposite directions. Triplets are sampled with the distance weighted sampling strategy when training the student, and with the focal weighted sampling strategy when training the generator. The focal weighted sampling tends to the negative neither too far nor too close, thus reserving global data diversity. Then, the embeddings of the student are optimized in opposite directions in terms of the triplets extracted by the distance weighted sampling.}
   \label{fig:framework}
\end{figure*}

\noindent\textbf{Overviews.} Let $\mathcal{T}$ be the teacher pre-trained on dataset $\mathcal{X}$ with well-optimized parameters $\theta_{t}$. $\mathcal{T}$ maps the input sample $x\in\mathcal{X}$ to a latent representation $\boldsymbol{e}_{t}=F_{t}(x)\in\mathbb{R}^{D_{t}}$ by the feature extractor $F_t$. We aim to teach a new student model parameterized by weights $\theta_{s}$ that still works well on the dataset $\mathcal{X}$ without access to the real training data. The latent representation of sample $x$ in the student is $\boldsymbol{e}_{s}=F_{s}(x)\in\mathbb{R}^{D_{s}}$. To train the student with synthetic data, we consider two phases in data-free KD. First, a generator $\mathcal{G}$ with parameters $\theta_{g}$ is trained based on the teacher with loss $\mathcal{L}_{g}$, which takes a random vector $\textbf{z}\in\mathbb{R}^{D_{z}}$  with dimension $D_{z}$ as input and outputs synthetic sample $x^{\prime}=\mathcal{G}(z)$ to construct the synthetic sample set $\mathcal{X}^{\prime}$. The class label $y^{\prime}$ of $x^{\prime}$ is given by the teacher, i.e. $y^{\prime}=\arg max(\boldsymbol{p}_{t}), \boldsymbol{p}_{t}=C(\boldsymbol{e}_{t})$ where $C$ is the classifier head of the teacher. Second, batches of samples are extracted from $\mathcal{X}^{\prime}$ for training the student based on the student learning loss $\mathcal{L}_{s}$. The two phases are performed alternatively to obtain the well-optimized parameter $\theta_{s}$. The overall pipeline of our proposed RGAL with these two phases is shown in Figure \ref{fig:framework}.

In Section \ref{sec:method}, we re-design the generator loss $\mathcal{L}_{g}$ for image synthesis and the student loss $\mathcal{L}_{s}$ for training the student model, where the embedding distances of synthetic samples are optimized adversarially with two opposite triplet losses. Thus, the intra-class diversity and inter-class confusion of the samples can be effectively guaranteed. Meanwhile, it improves the discrimination ability of the student for different classes. Then a focal weighted sampling strategy is introduced in Section \ref{sec:sampling} for selectively getting samples of triplets in terms of relative embedding distances. This resolves the potential optimization conflicts due to plain negative sampling, which reduces the impact of inter-class confusion on global sample diversity. The implementation details of RGAL are presented in Section \ref{sec:detail}.

\subsection{Relation-guided Adversarial Learning}
\label{sec:method}

We first consider the image synthesis phase, which aims for generating samples with intra-class diversity and inter-class confusion. To separately optimize the samples of the same class and those of different classes, we first need to refer to represent the relation among samples in terms of predicted class labels by the teacher. Typically, we randomly select one sample as the anchor. Then we define that the positive has the same label as the anchor, whereas the negative has a different label from the anchor. Thus the sample relation can be represented by an embedding triplet $(\boldsymbol{e}^{a}_{s}, \boldsymbol{e}^{p}_{s}, \boldsymbol{e}^{n}_{s})$ with a chosen anchor embedding $\boldsymbol{e}^{a}_{s}$. The embeddings of the positive and the negative are denoted by $\boldsymbol{e}^{p}_{s}$ and $\boldsymbol{e}^{n}_{s}$, respectively.

Using the sampled set of triplets, relation-guided adversarial learning is performed by incorporating additional adversarial triplet losses in the embedding space. We suppose an opposite triplet loss to perform two essential functions. Firstly, it is supposed to encourage $\boldsymbol{e}^{a}_{s}$ and $\boldsymbol{e}^{n}_{s}$ to be close to each other to ensure that the samples of different classes are enough inter-class confusing and hard to be distinguished. Secondly, it is supposed to encourage $\boldsymbol{e}^{a}_{s}$ and $\boldsymbol{e}^{p}_{s}$ to be far away from each other to ensure high diversity of samples of the same class. Thus, we notice that the optimization objective of the generator is just opposite to the optimization direction of commonly used triplet loss, which is given by:
\begin{equation}
  \begin{aligned}
  \mathcal{L}_{n t r i}= \big[\|\boldsymbol{e}^{a}_{s}-\boldsymbol{e}^{ n}_{s}\|^{2}-\|\boldsymbol{e}^{a}_{s}-\boldsymbol{e}^{p}_{s}\|^{2}+\tau\big]_{+}.
  \label{eq:n-Triplet-loss}
  \end{aligned}
\end{equation}
Therefore, the overall objective for image synthesis and training the generator can be formulated as follows:
\begin{equation}
  \begin{aligned}
  \mathcal{L}_{g}= - \mathcal{L}_{a d v} + \mathcal{L}_{n t r i},
  \label{eq:Generator-loss}
  \end{aligned}
\end{equation}
\noindent where $[\cdot]_{+}=\max(\cdot, 0)$ and $\|\cdot\|$ denotes the Euclidean distance. $\mathcal{L}_{a d v}=\sum_{i} \boldsymbol{p}_{t}^{(i)} \log \left(\boldsymbol{p}_{t}^{(i)} / \boldsymbol{p}_{s}^{(i)}\right)$ enlarge the output distance of the teacher and the student with a Kullback–Leibler divergence as in \citep{Micaelli2019ZeroShotKT}. In the image synthesis phase, maximizing $\mathcal{L}_{a d v}$ ensures that the generator explores the hard samples with the significant mismatch between the outputs of the student and the teacher. Besides, minimizing $\mathcal{L}_{n t r i}$ ensures that the generated samples are both highly inter-class confusing and intra-class diverse.

In the student training phase, the student is expected to be trained with stronger class-level discrimination. Therefore, the positive triplet loss for training the student can be defined as minimizing the distance between the anchor and the positive and maximizing the distance between the anchor and the negative. This is just the opposite of Equation \eqref{eq:n-Triplet-loss} and can be formulated as follows:
\begin{equation}
  \begin{aligned}
  \mathcal{L}_{t r i}= \big[\|\boldsymbol{e}^{a}_{s}-\boldsymbol{e}^{ p}_{s}\|^{2}-\|\boldsymbol{e}^{a}_{s}-\boldsymbol{e}^{n}_{s}\|^{2}+\tau\big]_{+},
  \label{eq:Triplet-loss}
  \end{aligned}
\end{equation}
\noindent where $\tau$ is a margin enforced between positive pairs and negative pairs for all possible triplets within a batch as in \citep{wu2017sampling} and \citep{zhang2022individual}. It keeps the distance difference greater than $\tau$, which enforces a margin between each pair of samples from one sample to all others with different labels. Then, the total loss for training the student can be formulated as follows:
\begin{equation}
  \begin{aligned}
  \mathcal{L}_{s}= \mathcal{L}_{a d v} + \mathcal{L}_{t r i},
  \label{eq:Student-loss}
  \end{aligned}
\end{equation}
\noindent where $\mathcal{L}_{a d v}$ is used to match the output distributions of the teacher and the student. $\mathcal{L}_{a d v}$ enables the student to generate predictions similar to the teacher. Furthermore, minimizing $\mathcal{L}_{t r i}$ forces the student to cluster samples of the same classes and distinguish samples of different classes.

\begin{figure}[t]
    \centering 
    \subfigure[Distance Weighted Sampling]{ 
    \begin{minipage}{4cm}
    \centering   
    \includegraphics[width=1.0\textwidth]{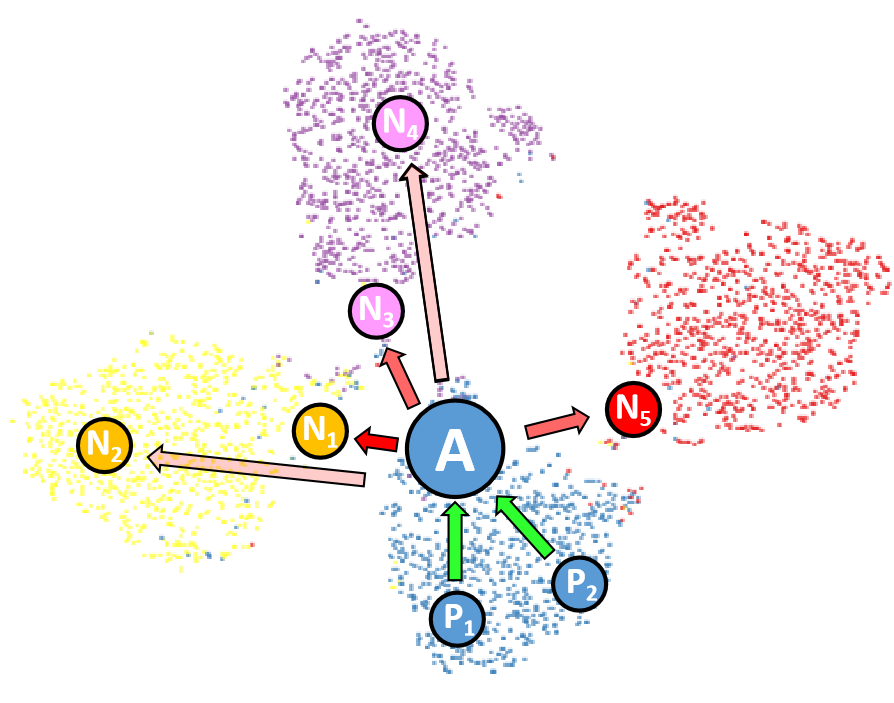}
    \end{minipage}
    }
    \subfigure[Focal Weighted Sampling]{ 
    \begin{minipage}{4cm}
    \centering 
    \includegraphics[width=1.0\textwidth]{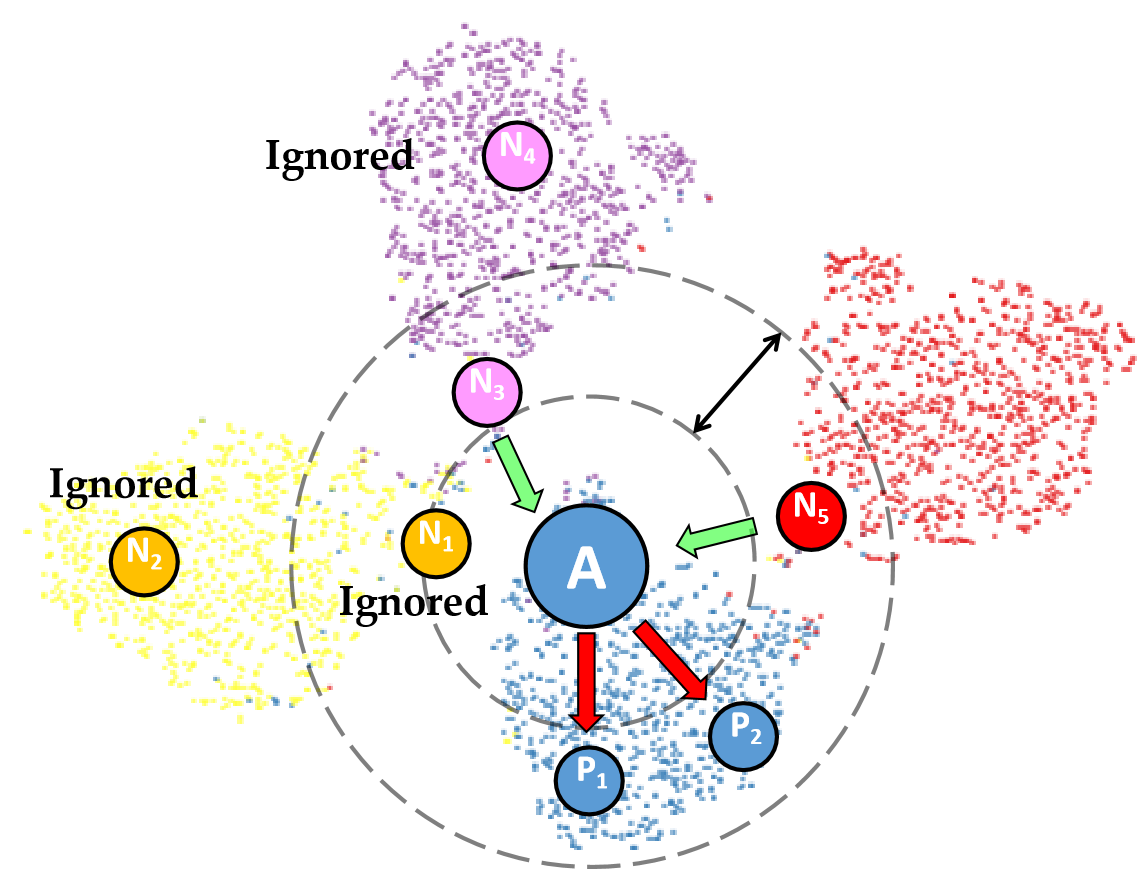}
    \end{minipage}
    }
    \caption{Illustration of different sampling strategies and optimization directions of the positive and the negative. Red denotes pushing away and green denotes pulling close, where color opacity denotes sampling probability.}
    \label{fig:sampling-example}  
\end{figure}

\begin{figure*}[t]
   \centering
   \includegraphics[width=0.95 \textwidth]{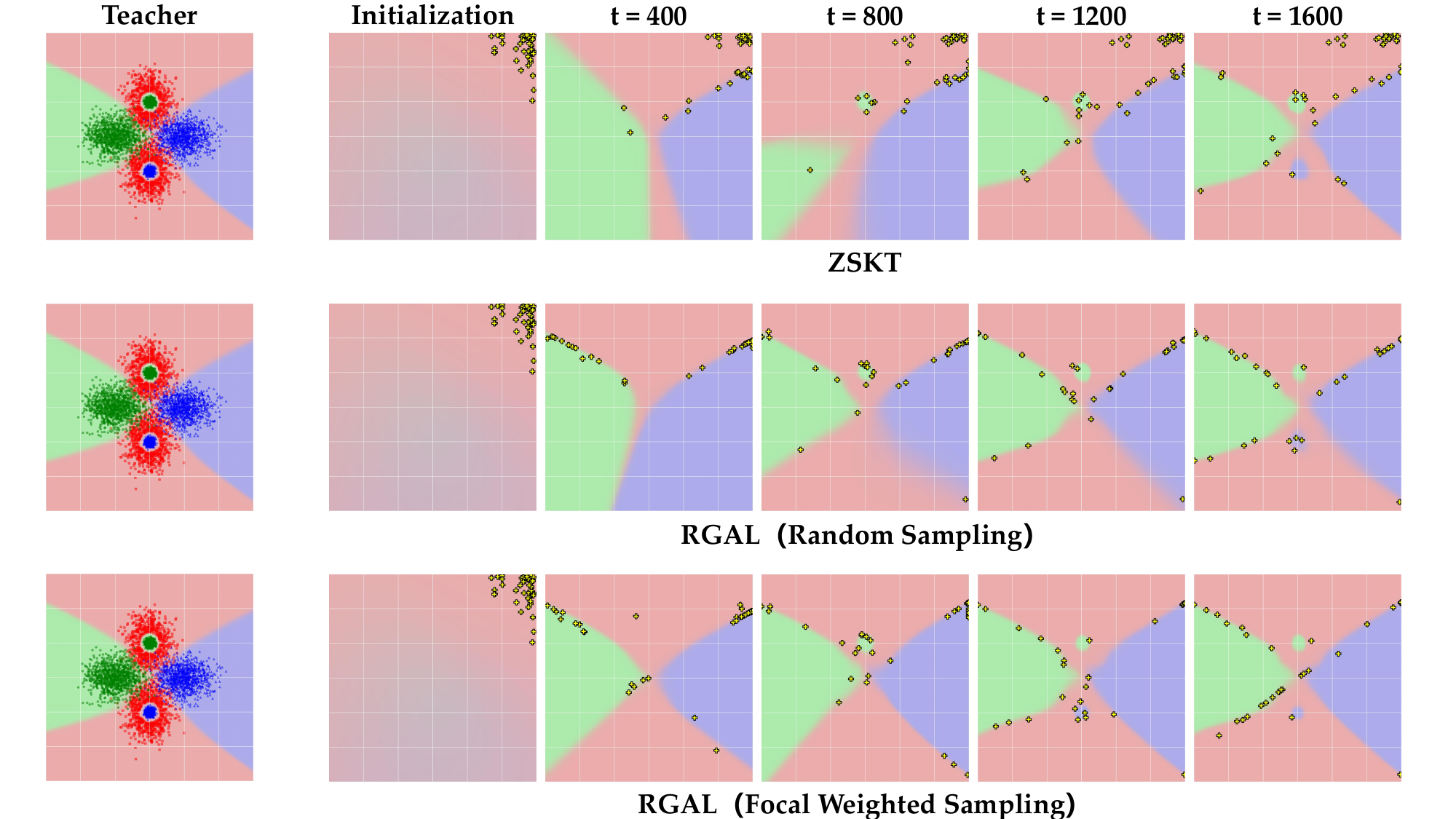} 
   \caption{Visualization of samples in the embedding space on a typical three-class problem from ZSKT \citep{Micaelli2019ZeroShotKT} and our methods. Pseudo points are randomly initialized away from the data manifold. The first line shows the result of ZSKT, in which the proposed adversarial formulation is widely adopted by the recent state-of-the-art methods. Both RGAL with adversarial triplet loss eliminate dense sample clusters and samples are more widely distributed.}
   \label{fig:decision-boundary}
\end{figure*}

\subsection{Focal Weighted Sampling Strategy}
\label{sec:sampling}
The triplet losses in Equations \eqref{eq:n-Triplet-loss} and \eqref{eq:Triplet-loss} both require sample relation in terms of triplets. Therefore, how to choose the samples for the triplet becomes a critical factor. One triplet includes three samples, i.e., a chosen anchor, a positive with the same label as the anchor, and a negative with a different label from the anchor. We first select an anchor from a batch and then select the positive with the same label as the anchor. When these two samples are fixed, the simplest method to select the negative is a random sampling strategy. Random sampling gets the negative for the triplet randomly from the remaining samples with different labels from a single batch. However, samples far away from the anchor are selected with equal probability as those close to the anchor, leading to suboptimal options for the negative. 

To be specific, in the student training phase, the samples of the same classes are expected to be closer together, while samples of different classes are expected to be farther apart. Therefore, there is little performance benefit from pushing away the negative that is far away because they have already been well distinguished in the embedding space. Only sampling negative samples close to the anchor also causes another issue a small disagreement may not always correspond to valuable samples because they are more likely just some outliers misclassified by the teacher. Therefore, we adopt a distanced weighted sampling strategy inspired by \citep{wu2017sampling} in the student learning phase. We sample the negative uniformly based on the relative distance with weights proportional to $1/f(d)$ where $d=\|\boldsymbol{p}_{t}^{a}-\boldsymbol{p}_{t}^{n}\|^{2}$. In this case, the distribution of pairwise distances follows:
\begin{equation}
  f(d) \propto d^{c-2}\left[1-\frac{1}{4} d^{2}\right]^{\frac{c-3}{2}}.
  \label{eq:distance}
\end{equation}
We also clip the sampling weights by $\lambda$ to avoid outliers as in \citep{wu2017sampling}. In other words, samples that have inverse distances within $\lambda$ are equally likely to be sampled as negative samples. Formally, given an anchor $\boldsymbol{e}^{a}_{s}$, distance weighted sampling samples $\boldsymbol{e}^{n}_{s}$ with probability as follows:
\begin{equation}
  \operatorname{Pr}\left(\boldsymbol{e}^{n}_{s} \mid \boldsymbol{e}^{a}_{s}\right) \propto \min \left(\lambda, \frac{1}{f\left(\|\boldsymbol{p}_{t}^{a}-\boldsymbol{p}_{t}^{n}\|^{2}\right)}\right),
  \label{eq:negative-pair}
\end{equation}
where we set $\lambda=0.5$ following the setup in \citep{wu2017sampling}. It provides more valuable triplets from the batch of samples, ensuring the samples close and with different labels are more likely to be pushed away. 

In the image synthesis phase, it is not necessary to pull together samples of different labels that are too close, because they have already been confusing enough. Besides, there is no need to pull close samples that are too far, because they are different enough to keep high data diversity. As shown in Figure \ref{fig:sampling-example}, for example, pulling close the anchor $A$ and sample $N_2$ and $N_4$ damages the global data diversity as they are different enough. Meanwhile, pulling close $A$ and $N_1$ is also not necessary as they are similar enough to keep inter-class confusion. Thus, we propose to focus on the samples only in local areas and select the negative for the triplet in terms of inverse distance weights:
\begin{equation}
 \operatorname{Pr}\left(\boldsymbol{e}^{n}_{s} \mid \boldsymbol{e}^{a}_{s}\right) \propto\left\{\begin{array}{cc}
 \frac{1}{f\left(\|\boldsymbol{p}_{t}^{a}-\boldsymbol{p}_{t}^{n}\|^{2}\right)} & \text { if } \lambda_{l}<1/f(d)<\lambda_{u}\\
 0 &  \text{otherwise}
\end{array}\right.,
\label{eq:focal-sampling}
\end{equation}
where we set $\lambda_{l}=0.4$ and $\lambda_{u}=1.0$ by default. We will show the effect of the two parameters on model performance in the ablation studies in Section \ref{sec:ablation}.

To more intuitively demonstrate the effectiveness of our proposed method, we conduct a three-class toy experiment following the one in ZSKT \citep{Micaelli2019ZeroShotKT}, which proposes the primary data-free knowledge distillation paradigms with an adversarial manner. We do not include other recent methods as they follow the same adversarial learning strategy. We show the dynamic samples and the learned student in ZSKT and our proposed adversarial learning pipeline with and without focal weighted sampling strategy, as shown in Figure \ref{fig:decision-boundary}. Our proposed adversarial learning pipeline avoids clustered samples and the proposed focal weighted sampling provides faster convergence of student learning. 

Since most random samples in a batch have different labels, the number of triplets from the batch is decided by the number of paired samples of the same classes. The batch sampling strategy determines the samples in each batch. However, the commonly used random sampling extracts a batch of samples randomly from the training data. The classes of the samples are too spread out, resulting in a limited number of triplets. FaceNet \citep{schroff2015facenet} uses a batch size of 1800 to ensure the number of triplets, which is computationally expensive. To this end, a balancing strategy is used in RGAL to ensure that a sufficient number of triplets are sampled within a limited batch size. Assuming that the batch size is fixed, the first half of samples are obtained by random sampling as the anchor, while the last half of samples are guaranteed to have the same labels as the former half. A sufficient number of triplets are provided by paired sampling where $50\%$ of samples are guaranteed to be the anchor and the positive, respectively, and then the negative can be sampled unevenly in terms of Equation \eqref{eq:focal-sampling}. 

\begin{algorithm} 
	\caption{RGAL for data-free knowledge transfer} 
	\label{algorithm-framework} 
	\begin{algorithmic}
		\renewcommand{\algorithmicensure}{\textbf{Pretrain:}} 
		\renewcommand{\algorithmicrequire}{\textbf{Initialize:}}
		\renewcommand{\algorithmicfor}{\textbf{For}} 
		\renewcommand{\algorithmicend}{\textbf{End}} 
		\ENSURE $\mathcal{T}(\cdot;\theta_{t})$ 
		\REQUIRE $\mathcal{S}(\cdot;\theta_{s})$ 
		\STATE $\mathcal{B}\leftarrow\varnothing$
        \FOR{Each $i\in [0, N)$}
            \STATE $\textbf{z}\sim\mathcal{N}(0, I)$
            \STATE Initialize $\mathcal{G}(\cdot;\theta_{g})$;
            \FOR{each $j\in [0, N_{g})$}
                \STATE $x\leftarrow \mathcal{G}(\textbf{z};\theta_{g})$
                \STATE $\boldsymbol{p}_{t}, \boldsymbol{e}_{t}\leftarrow \mathcal{T}(x;\theta_{t})$
                \STATE $\boldsymbol{p}_{s}, \boldsymbol{e}_{s}\leftarrow \mathcal{S}(x;\theta_{s})$
                \STATE $[(\boldsymbol{e}_{s}^{a}, \boldsymbol{e}_{s}^{p}, \boldsymbol{e}_{s}^{n})]^{N_{t}} \leftarrow\operatorname{focal\,weighted\,sampling}(x, \boldsymbol{p}_{t})$;
                \STATE $\mathcal{L}_{g}\leftarrow\mathcal{L}_{n t r i}(\boldsymbol{e}_{s}^{a}, \boldsymbol{e}_{s}^{p}, \boldsymbol{e}_{s}^{n}) + \mathcal{L}_{others}(x)$
                \STATE $\theta_{g} \leftarrow \theta_{g}-\eta_{g} \cdot \frac{\partial L_{g}}{\partial \theta_{g}}$
            \ENDFOR
            
            $\mathcal{B}\leftarrow\mathcal{B}\cup x$
            \FOR{Each $j\in [0, N_{s})$}
                \STATE $x\leftarrow\operatorname{paired\,sampling}(\mathcal{B})$
                \STATE $\boldsymbol{p}_{t}, \boldsymbol{e}_{t}\leftarrow \mathcal{T}(x;\theta_{t})$
                \STATE $\boldsymbol{p}_{s}, \boldsymbol{e}_{s}\leftarrow \mathcal{S}(x;\theta_{s})$
                \STATE $[(\boldsymbol{e}_{s}^{a}, \boldsymbol{e}_{s}^{p}, \boldsymbol{e}_{s}^{n})]^{N_{t}} \leftarrow\operatorname{distance\,weighted\,sampling}(x, \boldsymbol{p}_{t})$;
                \STATE $\mathcal{L}_{s}\leftarrow \mathcal{L}_{t r i}(\boldsymbol{e}_{s}^{a}, \boldsymbol{e}_{s}^{p}, \boldsymbol{e}_{s}^{n}) + \mathcal{L}_{others}(x)$
                \STATE $\theta_{s} \leftarrow \theta_{s}-\eta_{s} \cdot \frac{\partial L_{s}}{\partial \theta_{s}}$
            \ENDFOR
        \ENDFOR
	\end{algorithmic} 
\end{algorithm}

\subsection{Learning Details}
\label{sec:detail}

The detailed learning process of our proposed RGAL is provided in Algorithm \ref{algorithm-framework}. After obtaining the well-trained teacher model, we randomly initialize the student model. Then, we alternately perform the adversarial training process in the image synthesis phase and the student training phase. The generator uses $\mathcal{L}_{g}$ to train to generate inter-class confusing and intra-class diverse sample set $\mathcal{X}^{\prime}$, while the student is then trained on samples from $\mathcal{X}^{\prime}$ with $\mathcal{L}_{s}$ to conduct the knowledge transfer process based on the generated samples.

In the image synthesis phase, minimizing Equation \eqref{eq:Generator-loss} allows hard samples to be sufficiently class-wise confusing and keep higher intra-class diversity. Meanwhile, to keep the high fidelity and realism of the generated samples, we add one-hot loss $\mathcal{L}_{o h}$ proposed by Chen et al. \citep{chen2019data}. It ensures that the outputs of the generated samples are close to the one-hot vector. BN regularization $\mathcal{L}_{b n}$ \citep{yin2020dreaming} is also widely used to generate images with high-quality reproduction by minimizing the divergence between the feature statistics and Batch Normalization. These losses are also used in the state-of-the-art methods \citep{choi2020data, fang2021contrastive}. Thus, the total loss for image synthesis in Equation \eqref{eq:Generator-loss} can be re-formalized as follows:
\begin{equation}
  \begin{aligned}
  \mathcal{L}_{g} = - \mathcal{L}_{a d v} + \mathcal{L}_{n t r i} + \mathcal{L}_{o h} + \mathcal{L}_{b n}.
  \label{eq:training-generator}
  \end{aligned}
\end{equation}

In the student training phase, the intermediate features are also important for knowledge transfer. Inspired by \citep{park2019relational} training the student model to form the same relational structure with that of the teacher, matching the sample relation of the student and the teacher directly also provides performance benefit \citep{wu2020learning}. 

However, there is still a margin between the teacher and the student since they may have a different number of channels. It prevents the student from finding its own optimization space if we directly match the embeddings between the student and the teacher. A fully connected layer $\operatorname{F c}:\mathbb{R}^{D_s}\rightarrow\mathbb{R}^{D_t}$ is used to solve this problem:
\begin{equation}
   \mathcal{L}_{e m b}=\|\boldsymbol{e}_{t} - \operatorname{F c}(\boldsymbol{e}_{s})\|^{2}.
  \label{eq:embedding-loss}
\end{equation}
\noindent Therefore, the total loss in Equation \eqref{eq:Student-loss} for knowledge transfer is redefined as follows:
\begin{equation}
    \mathcal{L}_{s}=\mathcal{L}_{a d v}+\mathcal{L}_{t r i}+\mathcal{L}_{e m b}.
    \label{eq:Distillation-loss}
\end{equation}
As suggested by \citep{fang2021contrastive}, at the beginning of each epoch, we re-initialize the generator $\mathcal{G}$ and random input $\textbf{z}$ to synthesize new samples more easily distinguished from the historical ones. Moreover, a single batch is insufficient for training the student. To better alleviate the mode collapse issue, we build a data pool $\mathcal{B}$, which also participates in the training process by restoring samples with the lowest loss $\mathcal{L}_{g}$. We focus on samples of a single batch and initialize the $\mathcal{B}$ to $\varnothing$, avoiding a large amount of pre-synthesized samples in advance for $\mathcal{B}$, which is required by the state-of-the-art method \citep{fang2021contrastive}. 

\section{Experiments}

In this section, we present a comprehensive evaluation of our proposed method. We first demonstrate the performance of our method on several benchmark datasets, including CIFAR-10, CIFAR-100, Tiny-ImageNet, and ImageNet, utilizing various teacher-student model configurations. We then conduct an ablation study to analyze the contributions of different components of our method, particularly focusing on the impacts of specific loss functions and sampling strategies. This study helps to elucidate the importance of each component in enhancing model performance. Additionally, we compare our method against state-of-the-art data-free knowledge distillation techniques to establish its relative effectiveness. Furthermore, we include visualization analyses to illustrate the diversity and quality of the generated synthetic samples, as well as the distribution of their embeddings in the feature space. Through these extensive experiments, we aim to thoroughly validate the robustness and efficiency of our method in achieving superior performance in data-free knowledge distillation.

\subsection{Datasets and Settings}

\noindent\textbf{Datasets}. We evaluate existing data-free knowledge distillation methodes and our method on CIFAR-10, CIFAR-100 \citep{krizhevsky2009learning}, and Tiny-ImageNet \citep{wu2017tiny}. These datasets are usually used for knowledge distillation. CIFAR-10 and CIFAR-100 contain 50,000 samples for training and 10,000 samples for testing, with a resolution of $32 \times 32$. Tiny-ImageNet consists of 100,000 samples for training and 10,000 samples for validation, whose resolution is $64 \times 64$. We further show how our method performs in the following extended experiments on the ImageNet \citep{deng2009imagenet}, which contains images with a resolution of $224\times224$.

\begin{table*}
    \centering
    \caption{Comparison experiments on various benchmarks. ``-" indicates no results reported in the paper.}
    \begin{tabular}{@{}lcccccccc@{}}
    \midrule
        Dataset & Teacher & Student                      & ZSKT & DAFL  & DFQ   & DeepInv & CMI & RGAL \\ \midrule
        \multirow{4}{*}{CIFAR-10} & resnet34 & resnet18  & 93.32 & 93.21 & 94.61 & 93.26  & 94.84 & \textbf{95.08} \\ 
        ~ & wrn40-2 & wrn16-1                            & 83.74 & 77.83 & 86.14 & 83.04  &  90.01 & \textbf{91.14} \\ 
        ~ & wrn40-2 & wrn40-1                            & 86.07 & 81.33 & 91.69 & 86.85  & 92.78 & \textbf{92.84} \\ 
        ~ & wrn40-2 & wrn16-2                            & 89.66 & 81.55 & 92.01 & 89.72  & 92.52 & \textbf{92.67} \\ \midrule
        \multirow{5}{*}{CIFAR-100} & resnet34 & resnet18 & 67.74 & 75.52 & 77.01 & 61.32  &  \textbf{77.04} & 77.01  \\ 
        ~ & wrn40-2 & wrn16-1                            & 30.15 & 22.50 & 54.77 & 53.77  & 57.91 & \textbf{58.73} \\ 
        ~ & wrn40-2 & wrn40-1                            & 29.73 & 34.66 & 62.92 & 61.33  &  68.88 & \textbf{69.25}  \\ 
        ~ & wrn40-2 & wrn16-2                            & 28.44 & 40.00 & 59.01 & 61.34  &  68.75 & \textbf{69.71} \\ \midrule
        Tiny-ImageNet & resnet34 & resnet18              & - & -     & 63.73 & 62.38  & 64.01 & \textbf{64.23} \\ \midrule
    \end{tabular}
    \label{table:Comparison}
\end{table*}

\noindent\textbf{Generator Architecture}. As in \citep{fang2021contrastive}, the size $D_z$ of the generator input $\textbf{z}$ is $256$ for CIFAR and $512$ for Tiny-ImageNet. $\textbf{z}$ is sampled from the standard normal distribution. We train a single generator for distillation, where the generator is trained to produce all classes. The generator architecture is set up following \citep{fang2021contrastive} and is illustrated as follows:
$$\begin{array}{c}
\midrule \textbf{z} \in \mathbb{R}^{D z} \sim \mathcal{N}(0, I) \\
\midrule \text { Linear(Dz)} \rightarrow \text{(w / 4)} \times \text{(h / 4)} \times \text{(Dz / 2)} \\
\midrule \text { Reshape, BN } \\
\midrule \text { Upsample } \times 2 \\
\midrule 3 \times 3 \text { Conv(Dz) } \rightarrow \text { Dz }, \text { BN, LeakyReLU } \\
\midrule \text { Upsample } \times 2 \\
\midrule 3 \times 3 \text { Conv(Dz) } \rightarrow \text { Dz / 2}, \text { BN, LeakyReLU } \\
\midrule 3 \times 3 \text { Conv(Dz / 2) } \rightarrow 3 \text {, Sigmoid } \\
\midrule
\end{array}$$
 
\noindent\textbf{Network architecture}. As in \citep{fang2021contrastive}, we focus on two network architectures, e.g., resnet \citep{he2016deep} and Wide resnet (WRN) \citep{zagoruyko2016wide} with a different number of blocks or channels. We modify the first convolutional layer for resnet on CIFAR-10 and CIFAR-100 with kernel size $3 \times 3$ instead of the kernel size $7 \times 7$ as ZSKT \citep{Micaelli2019ZeroShotKT}.

\noindent\textbf{Implementation Details}. For data-free knowledge distillation, we use stochastic gradient descent (SGD) optimizer with momentum 0.9 \citep{kingma2014adam} and weight decay $5e^{-4}$ for training. The initial learning rate is set to 0.1 for pre-training and distillation training for 200 epochs. Cosine annealing scheduling \citep{loshchilov2016sgdr} is used to set the learning rate of each parameter group. To evaluate our methods, we take the top-1 accuracy on the test set as our metric for classification tasks like \citep{yin2020dreaming, choi2020data, fang2021contrastive}.

\subsection{Comparison with State-of-the-arts}
\label{sec:dfkd}

We first evaluate our proposed RGAL on data-free knowledge distillation and compare it with ZSKT \citep{Micaelli2019ZeroShotKT}, DAFL \citep{chen2019data}, DeepInversion \citep{yin2020dreaming}, DFQ \citep{choi2020data}, and CMI \citep{fang2021contrastive}, which are state-of-the-art methods for data-free knowledge distillation. ZSKT uses adversarial losses for image synthesis and training the student, while DAFL adds one-hot losses for convincing image synthesis. DeepInv and DFQ add BN regularization based on DAFL. CMI further proposes a contrastive-learning-based method for diverse image synthesis. By default, the teacher model is pre-trained on the labeled datasets, while the student model is trained on synthetic samples and evaluated on real data.

We evaluate our method and the compared methods based on three different settings, which contains different teacher models, different student models, and different datasets with different numbers of classes. As shown in Table \ref{table:Comparison}, our method performs better than most methodes on most test benchmarks. Except for resnet18 on CIFAR-100, the accuracy of the student obtained by our method gains improvements of $+0.2\%$ to $+1.3\%$ higher than the state-of-the-art on CIFAR-10 and CIFAR-100 with a variety of student models. On the more difficult Tiny-Imagenet dataset, the student models of ZSKT and DAFL are difficult to converge, and the accuracy of our method on Tiny-Imagenet gains an improvement of $+0.22\%$ higher than that of the state-of-the-art.

Table \ref{tab:data-amount} also reports the pre-synthesized data amount required by the CMI and modified CMI with our proposed RGAL, which replaces the contrastive loss with the triplet losses proposed in RGAL. The results show that our method avoids a large amount of pre-synthesized data by comparing the samples in a single batch rather than previous data. Without the usage of pre-synthesized data, CMI being equipped with RGAL demonstrates much better performance with accuracy improvements of +1.07\% on resnet18 and +1.26\% on wrn16-1. The results show that RGAL improves data efficiency effectively while ensuring accuracy.

\begin{table}[t] \small
\centering
    \caption{Pre-synthesized data amount required by CMI and CMI with RGAL, which replaces the contrastive loss with losses from our RGAL on CIFAR-10.}
    \begin{tabular}{@{}lccc@{}}
    \midrule
    Student Model & Method & Data Amount & Top1-Acc \\ \midrule
    \multirow{3}{*}{resnet18}  & CMI & 51.2k & 94.84 \\ 
    ~  & CMI & No & 93.94 \\ 
    ~  & CMI+RGAL & No & \textbf{95.01} \\ \midrule
    \multirow{3}{*}{wrn16-1}  & CMI & 51.2k & 90.01 \\ 
    ~  & CMI & No & 88.82 \\
    ~  & CMI+RGAL & No & \textbf{90.08} \\ \midrule
    \end{tabular}%
  \label{tab:data-amount}
\end{table}

\begin{figure*}
   \centering
   \includegraphics[width=0.98\textwidth]{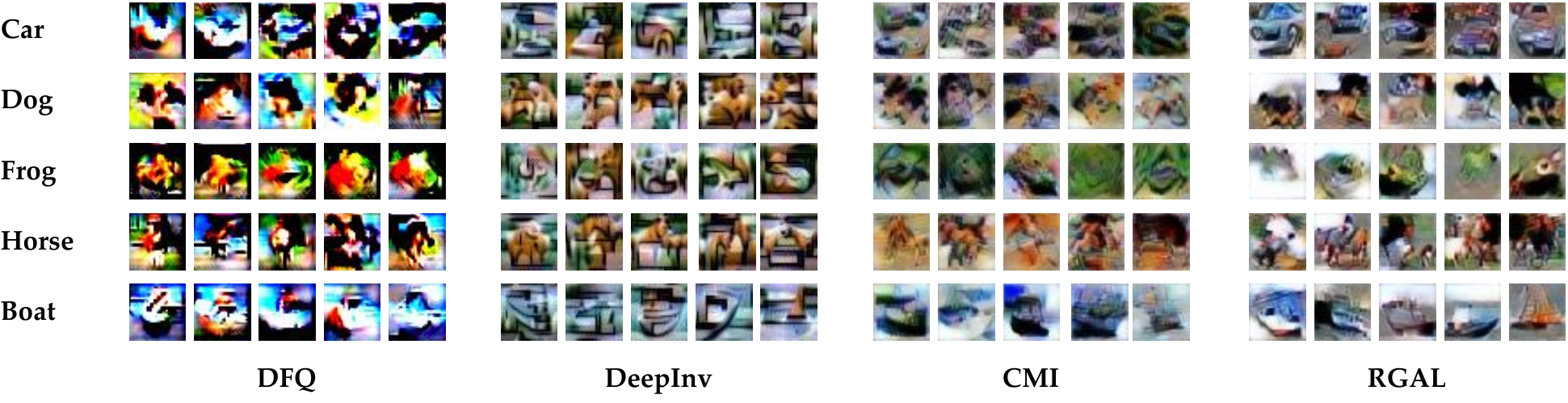} 
   \caption{A batch of generated samples for knowledge transfer on CIFAR-10 from a trained WRN40-2 model. Compared with the others, the samples from RGAL show high diversity and stronger inter-class confusion among the same batch. We compare our proposed RGAL with the state-of-the-arts.}
   \label{fig:visualization}
\end{figure*}

\begin{table}[t] \small
\centering
    \caption{Ablation study by cutting off different losses.}
    \begin{tabular}{@{}lccc@{}}
    \midrule
    Dataset & Method & resnet18 & wrn16-1 \\ \midrule
    \multirow{4}{*}{CIFAR-10} & RGAL & \textbf{95.08} & \textbf{91.14} \\ 
    ~ & w/o $\mathcal{L}_{n t r i}$ & 94.07 & 90.56 \\ 
    ~ & w/o $\mathcal{L}_{e m b}$ & 93.98 & 89.98 \\ 
    ~ & w/o $\mathcal{L}_{t r i}$ & 94.43 & 90.85 \\ \midrule 
    \multirow{4}{*}{CIFAR-100} & RGAL & \textbf{77.01} & \textbf{58.73} \\ 
    ~ & w/o $\mathcal{L}_{n t r i}$ & 76.49 & 57.45 \\ 
    ~ & w/o $\mathcal{L}_{e m b}$ & 76.80 & 58.40\\ 
    ~ & w/o $\mathcal{L}_{t r i}$ & 76.89 & 57.97 \\ \midrule
    \end{tabular}%
  \label{tab:ablation-loss}
\end{table}

\subsection{Visualization}

To visualize the synthetic samples in a single batch, we take the results from a trained WRN40-2 on CIFAR-10, as shown in Figure \ref{fig:visualization}. For visualization, we select the top-5 samples with the highest confidence of the five categories from all the samples in the steps of one image synthesis epoch. DFQ can hardly contain images with semantic information, while CMI and Deepinv suffer from data redundancy in the same batch of samples used for training. Our method achieves better visual quality and diversity in a single batch than the other methods. In addition, samples of different classes generated from RGAL are also more visually similar. For example, the samples of the car and the samples of the boat are visually similar, suggesting higher inter-class confusion.

\subsection{Discussion}
\label{sec:ablation}

\noindent\textbf{Loss Functions}. We first investigate the contribution of different losses proposed in our method, $\mathcal{L}_{n t r i}$, $\mathcal{L}_{e m b}$ and $\mathcal{L}_{t r i}$. Each loss is separately turned off to check its effectiveness. As shown in Table \ref{tab:ablation-loss}, turning off any of these individual losses leads to performance reduction, and the loss term $\mathcal{L}_{t r i}$ matters to better performance. The reason is that it guarantees both sample diversity and confusion. Moreover, it can be seen that adding $\mathcal{L}_{e m b}$ and $\mathcal{L}_{t r i}$ can also help to improve the performance of the student. Because $\mathcal{L}_{e m b}$ guides the student to learn the distribution of valuable patterns from the teacher and $\mathcal{L}_{t r i}$ helps to form stronger and more distinguishable embeddings among different classes.

\begin{table}[t] \small
\centering
  \caption{Effect of different batch sampling strategies.}
    \begin{tabular}{@{}lccc@{}}
    \midrule
    Dataset & Batch Sampling & resnet18 & wrn16-1 \\ \midrule
    \multirow{2}{*}{CIFAR-10} & random & 95.02 & 91.10 \\ 
    ~ & paired (Ours) & \textbf{95.08} & \textbf{91.14} \\ \midrule
    \multirow{2}{*}{CIFAR-100} & random & 76.91 & 58.45 \\ 
    ~ & paired (Ours) & \textbf{77.01} & \textbf{58.73} \\ \midrule
    \end{tabular}%
  \label{tab:batch-sample-strategy}
\end{table}

\noindent\textbf{Batch Sampling Strategy}. We also analyze the effectiveness of the batch sampling strategy to construct each batch, i.e., random sampling and paired sampling. The batch sampling strategy determines the samples in each batch. Random sampling extracts a batch of samples randomly from the training data, while paired sampling extracts a batch of samples with half of the paired labels. Thus, the samples in the batch are fixed. Then, the triplets are selected in this batch. Paired sampling ensures that the last half of samples are labeled the same as the first half of samples. Paired sampling provides a more adequate number of triplets for triplet loss compared to random sampling. As shown in Table \ref{tab:batch-sample-strategy}, paired sampling better improves the performance of our proposed RGAL than random sampling in various settings.

\noindent\textbf{Triplet Sampling Strategy}. We also show the effect of triplet sampling strategies in the image synthesis phase, i.e., focal weighted sampling, distance weighted sampling, and random sampling. We first select an anchor from the batch and then select the positive with the same label as the anchor. When these two samples are fixed, these three sampling strategies are designed to extract the negative. Table \ref{tab:sample-strategy} shows how these three sampling strategies used in the image synthesis phase affect the performance of RGAL. The accuracy of models with focal weighted sampling is better than the other two sampling strategies. The reason may be that focal weighted sampling ensures that the samples close enough or far away cannot be selected as negative samples. This prevents these negative samples from being pulled too close to the anchor, thus preserving global sample diversity. 

\begin{table}
  \centering
  \caption{Effect of different triplet sampling strategies.}
    \begin{tabular}{@{}lcccc@{}}
    \midrule
    Dataset & Sampling & resnet18 & wrn16-1 \\ \midrule
    \multirow{3}{*}{CIFAR-10} & random & 94.23 & 90.86 \\ 
    ~ & distance weighted & 95.02 & 91.03 \\ 
    ~ & focal weighted (Ours) & \textbf{95.08} & \textbf{91.14} \\ \midrule
    \multirow{3}{*}{CIFAR-100} & random & 76.16 & 58.49 \\ 
    ~ & distance weighted & 76.90 & 58.03 \\
    ~ & focal weighted (Ours) & \textbf{77.01} & \textbf{58.73} \\ \midrule
  \end{tabular}
  \label{tab:sample-strategy}
\end{table}

\begin{table}
  \centering
  \caption{Effect of different embedding representations.}
    \begin{tabular}{@{}lcccc@{}}
    \midrule
    Dataset & Logits & Global                   & resnet18        & wrn16-1 \\ \midrule
    \multirow{3}{*}{CIFAR-10}  & \checkmark &   & 95.06           & \textbf{91.41} \\ 
    ~  & & \checkmark                           & \textbf{95.08}  & 91.14 \\ 
    ~  & \checkmark & \checkmark                & 95.02           & 91.15 \\ \midrule
     \multirow{3}{*}{CIFAR-100} & \checkmark &  & 76.87           & 58.73  \\ 
    ~ & & \checkmark                            & \textbf{77.01}  & \textbf{58.91}  \\ 
    ~ & \checkmark & \checkmark                 &  76.91          & 58.71  \\ \midrule
  \end{tabular}
  \label{tab:embedding-representations-wrn}
\end{table}

\noindent\textbf{Embedding Representations}. We analyze the choices for our embedding representation in the student model for triplet loss, as shown in Table \ref{tab:embedding-representations-wrn}. The results consist of logits, global embeddings, and their combination. The logits are the outputs before softmax in the student model and the global embeddings indicate the feature after global pooling. We can see that RGAL using global embeddings outperforms the one using other representations in most cases. The reason may be that the logits of samples with different labels are so close that they only have limited class-specific content. Therefore, global embedding is more advantageous than logits for tasks with more classes. 

\begin{table}
  \centering
  \caption{Ablation study on hyper-parameter $\lambda$.}
    \begin{tabular}{@{}lccc@{}}
    \midrule
    Dataset & value of $\lambda$ & resnet18 & wrn16-1 \\ \midrule
    \multirow{5}{*}{CIFAR-10} & $\lambda=0.3$  & 94.35 & 91.10   \\
    ~ & $\lambda=0.4$                          & 94.81 & 91.13   \\
    ~ & $\lambda=\underline{0.5}$              & \textbf{95.08} & \textbf{91.14}   \\
    ~ & $\lambda=0.6$                          & 94.92 & 91.10   \\
    ~ & $\lambda=0.7$                          & 92.04 & 90.89   \\ \midrule 
    \multirow{5}{*}{CIFAR-100} & $\lambda=0.3$ & 76.89 & 58.67   \\
    ~ & $\lambda=0.4$                          & 76.97 & 58.65   \\
    ~ & $\lambda=\underline{0.5}$              & \textbf{77.01} & \textbf{58.73}   \\
    ~ & $\lambda=0.6$                          & 76.90 & 58.70   \\ 
    ~ & $\lambda=0.7$                          & 76.25 & 58.49   \\ \midrule 
  \end{tabular}
  \label{tab:ablation-lambda}
\end{table}

\begin{table}
  \centering
  \caption{Ablation study on hyper-parameter $\lambda_l$.}
    \begin{tabular}{@{}lccc@{}}
    \midrule
    Dataset & value of $\lambda$ & resnet18 & wrn16-1 \\ \midrule
    \multirow{5}{*}{CIFAR-10} & $\lambda=0.3$  & 95.00 & 90.76   \\
    ~ & $\lambda=0.4$                          & 95.04 & 90.89   \\
    ~ & $\lambda=\underline{0.5}$              & \textbf{95.08} & \textbf{91.14}   \\
    ~ & $\lambda=0.6$                          & 94.99 & 91.10   \\
    ~ & $\lambda=0.7$                          & 94.86 & 90.79   \\ \midrule 
    \multirow{5}{*}{CIFAR-100} & $\lambda=0.3$ & 69.91 & 58.70   \\
    ~ & $\lambda=0.4$                          & \textbf{77.05} & 58.70   \\
    ~ & $\lambda=\underline{0.5}$              & 77.01 & \textbf{58.73}   \\
    ~ & $\lambda=0.6$                          & 76.88 & 58.68   \\
    ~ & $\lambda=0.7$                          & 76.80 & 58.65   \\ \midrule 
  \end{tabular}
  \label{tab:ablation-lambda_l}
\end{table}

\begin{table}
  \centering
  \caption{Ablation study on hyper-parameter $\lambda_u$.}
    \begin{tabular}{@{}lccc@{}}
    \midrule
    Dataset & value of $\lambda_u$ & resnet18 & wrn16-1 \\ \midrule
    \multirow{5}{*}{CIFAR-10} & $\lambda_u=0.8$    & 94.26             & 90.08 \\
    ~ & $\lambda_u=0.9$                            & 94.49             & 90.44 \\
    ~ & $\lambda_u=\underline{1.0}$                & \textbf{95.08}    & 91.14 \\
    ~ & $\lambda_u=1.1$                            & 95.06             & 90.90 \\
    ~ & $\lambda_u=1.2$                            & 95.00             & \textbf{91.15} \\ \midrule 
    \multirow{5}{*}{CIFAR-100} & $\lambda_u=0.8$   & 69.88             & 58.51 \\
    ~ & $\lambda_u=0.9$                            & 69.94             & 58.50 \\
    ~ & $\lambda_u=\underline{1.0}$                & \textbf{77.01}    & 58.73 \\
    ~ & $\lambda_u=1.1$                            & \textbf{77.01}    & \textbf{58.75} \\
    ~ & $\lambda_u=1.2$                            & 76.89             & 58.71 \\ \midrule 
  \end{tabular}
  \label{tab:ablation-lambda_u}
\end{table}

\noindent\textbf{Hyper-parameters}. The proposed focal weighted sampling strategy introduces three hyper-parameters, i.e., $\lambda$, $\lambda_l$, and $\lambda_u$. Hand-picked values for these hyper-parameters may lead to suboptimal results. Therefore, we show more ablation studies about these hyper-parameters to show their effectiveness. A smaller $\lambda_l$ and a larger $\lambda_u$ indicate a wider range of choices for negative samples when calculating $\mathcal{L}_{n t r i}$. The results are shown in Tables \ref{tab:ablation-lambda}, \ref{tab:ablation-lambda_l}, and \ref{tab:ablation-lambda_u}. Typically, $\lambda=0.5$ gives the best performance. A smaller $\lambda_l$ and a larger $\lambda_u$ lead to better performance. With a small change within $\pm0.1$ in parameter values, the accuracy difference of the model is within $\pm0.20\%$, proving that our method is relatively hyper-parameter insensitive. 

\subsection{Analysis of Synthesized Images}

\begin{figure}[t]
    \centering
   \includegraphics[width=0.5\textwidth]{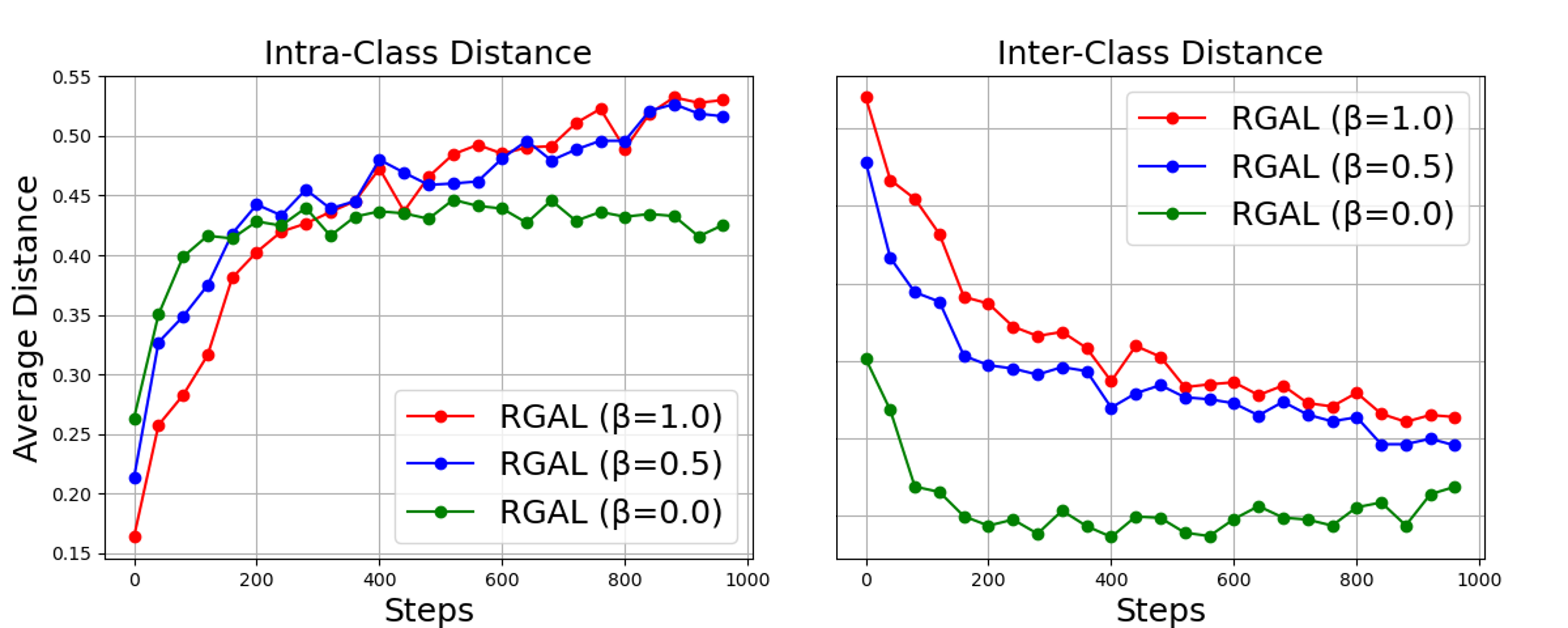}
   \caption{Average distances of samples of the same classes and samples of different classes. The samples are generated with the on class-conditional $224 \times 224$ images obtained by RGAL with different loss weights $\beta$ on $\mathcal{L}_{n t r i}$ given a resnet50 classifier pre-trained on ImageNet. $\beta=0$ indicates not applying $\mathcal{L}_{n t r i}$ for image synthesis.}
   \label{fig:l1-distance-imagenet}
\end{figure}

\begin{figure}[t]
    \centering
   \includegraphics[width=0.45\textwidth]{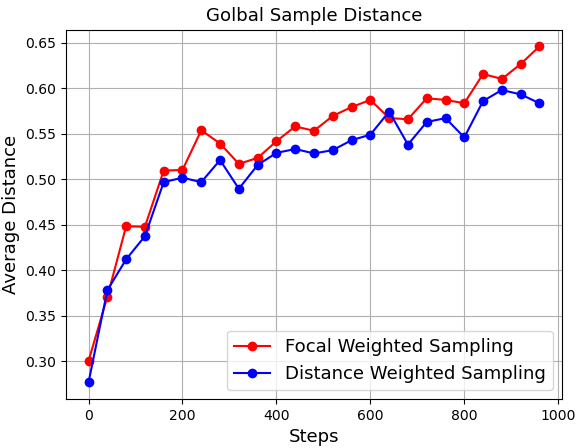}
   \caption{Average distances of global samples. These samples are generated on class-conditional $224 \times 224$ images obtained by RGAL with different sampling strategies given a resnet50 pre-trained on ImageNet.}
   \label{fig:global-distance-imagenet}
\end{figure}

\noindent\textbf{Data Diversity}. Given a set of data $\mathcal{X}$, an intuitive description of data diversity would be “how distinguishable are the samples from the dataset” \citep{fang2021contrastive}. Thus we can develop a clear definition of data diversity as the following:
\begin{equation}
    \mathcal{L}_{\text {div }}(\mathcal{X})=\mathbb{E}_{x_{i}, x_{j} \in \mathcal{X}}\left[d\left(x_{i}, x_{j}\right)\right],
\end{equation}
where $d$ is the $\ell_1$ distance. We refer to $\mathcal{L}_{\text {div }}$ as global diversity, which indicates the diversity of all samples.

As stated previously, we expect higher intra-class diversity and inter-class confusion in the synthetic samples. In order to quantitatively demonstrate the effects of RGAL on the above two indicators, we define intra-class diversity as the diversity of samples with the same labels:
\begin{equation}
    \mathcal{L}_{\text {intra-div }}(\mathcal{X})=\mathbb{E}_{x_{i}, x_{j} \in \mathcal{X}}\left[\mathbb{I}_{i j}^{\text {same }}d\left(x_{i}, x_{j}\right)\right],
    \label{eq:intra-class-diversity}
\end{equation}
where $\mathbb{I}_{i j}^{\text {same }}$ denotes whether $x_{i}$ and $x_{j}$ are of the same classes. Similarly, we can define inter-class confusion in the opposite of Equation \eqref{eq:intra-class-diversity} as follows:
\begin{equation}
    \mathcal{L}_{\text {inter-con }}(\mathcal{X})= 1 - \mathbb{E}_{x_{i}, x_{j} \in \mathcal{X}}\left[\mathbb{I}_{i j}^{\text {not-same }}d\left(x_{i}, x_{j}\right)\right],
\end{equation}
which indicates the similarity of the embeddings between each sample with different labels.

\begin{figure}[t]
    \centering  %居中
    \subfigure[Deepinv]{   %第一张子图
    \begin{minipage}{4cm}
    \centering    %子图居中
    \includegraphics[width=0.95\textwidth]{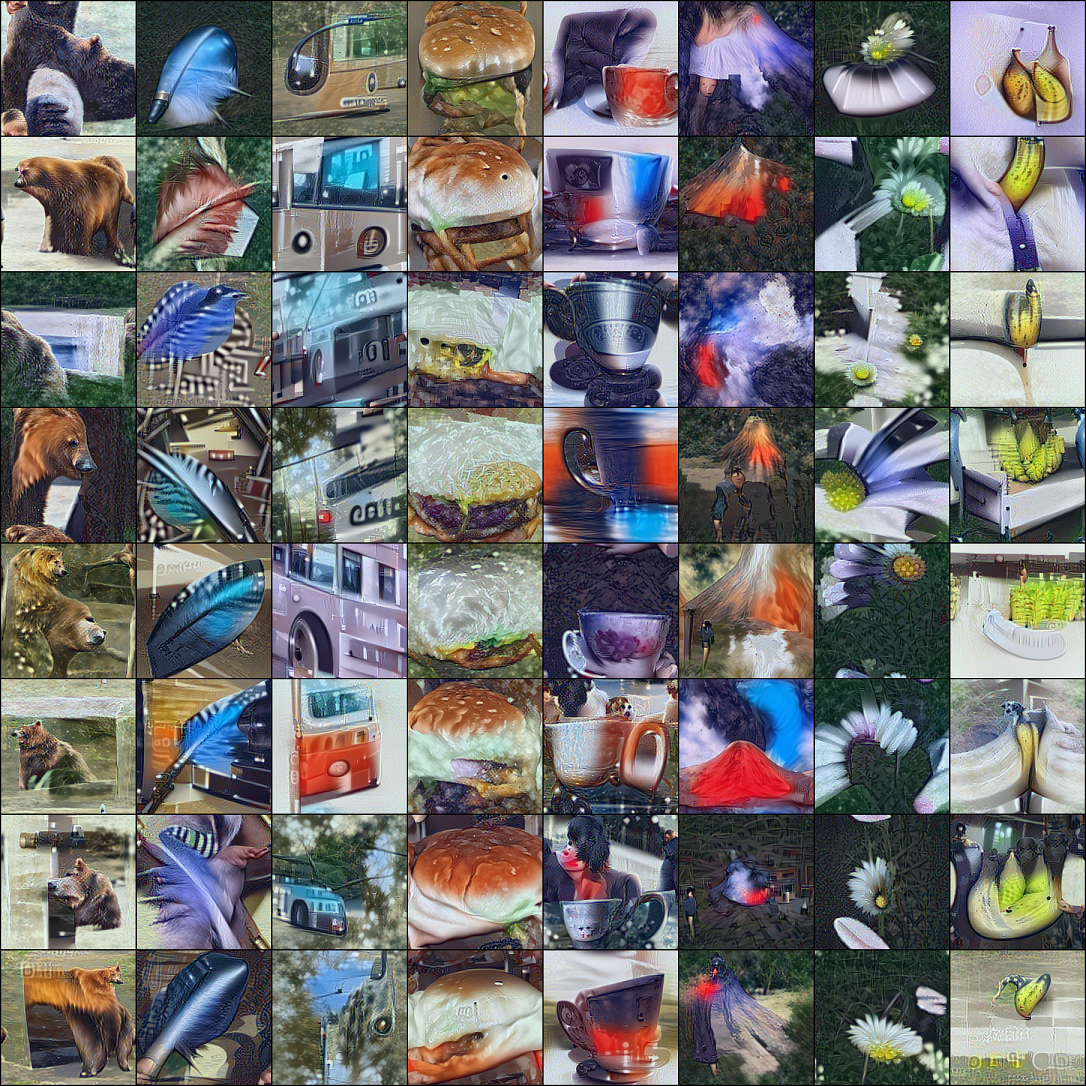}  %以pic.jpg的0.5倍大小输出
    \end{minipage}
    }
    \subfigure[Deepinv with $\mathcal{L}_{\text{ntri}}$ ($\beta=0.2$)]{ %第二张子图
    \begin{minipage}{4cm}
    \centering    %子图居中
    \includegraphics[width=0.95\textwidth]{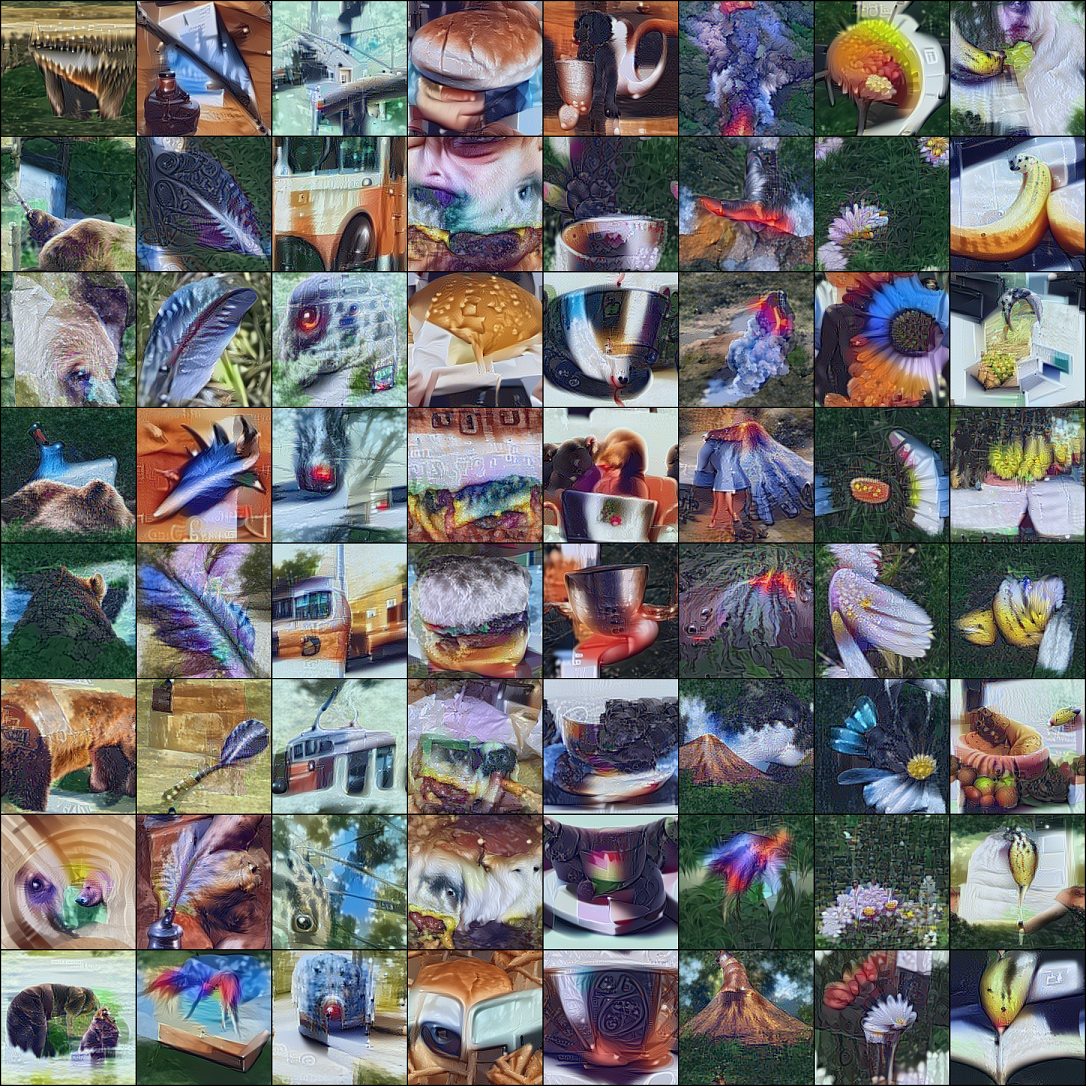}%以pic.jpg的0.5倍大小输出
    \end{minipage}
    }
    \subfigure[Deepinv with $\mathcal{L}_{\text{ntri}}$ ($\beta=0.5$)]{ %第二张子图
    \begin{minipage}{4cm}
    \centering    %子图居中
    \includegraphics[width=0.95\textwidth]{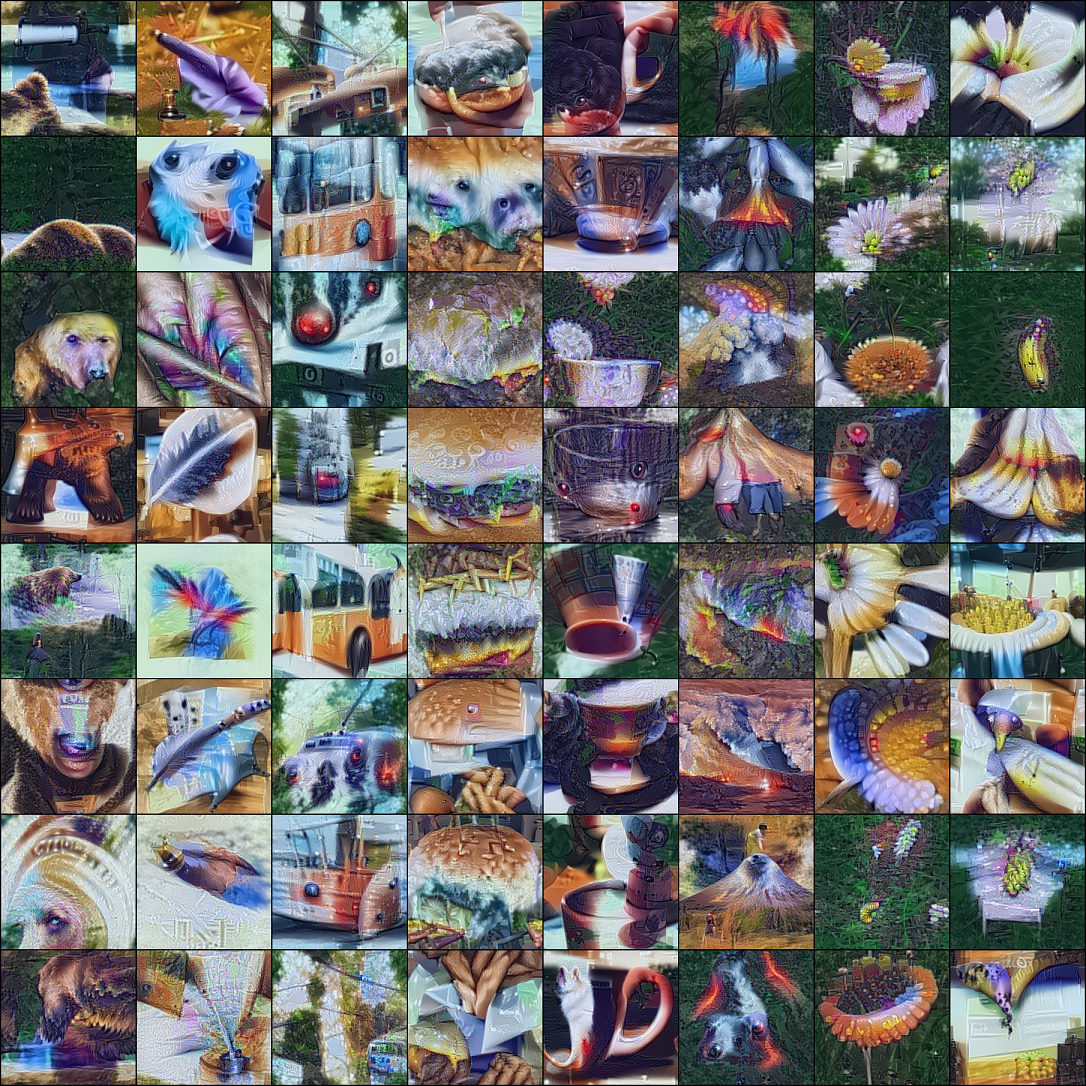}%以pic.jpg的0.5倍大小输出
    \end{minipage}
    }
    \subfigure[Deepinv with $\mathcal{L}_{\text{ntri}}$ ($\beta=1.0$)]{
    \begin{minipage}{4cm}
    \centering    %子图居中
    \includegraphics[width=0.95\textwidth]{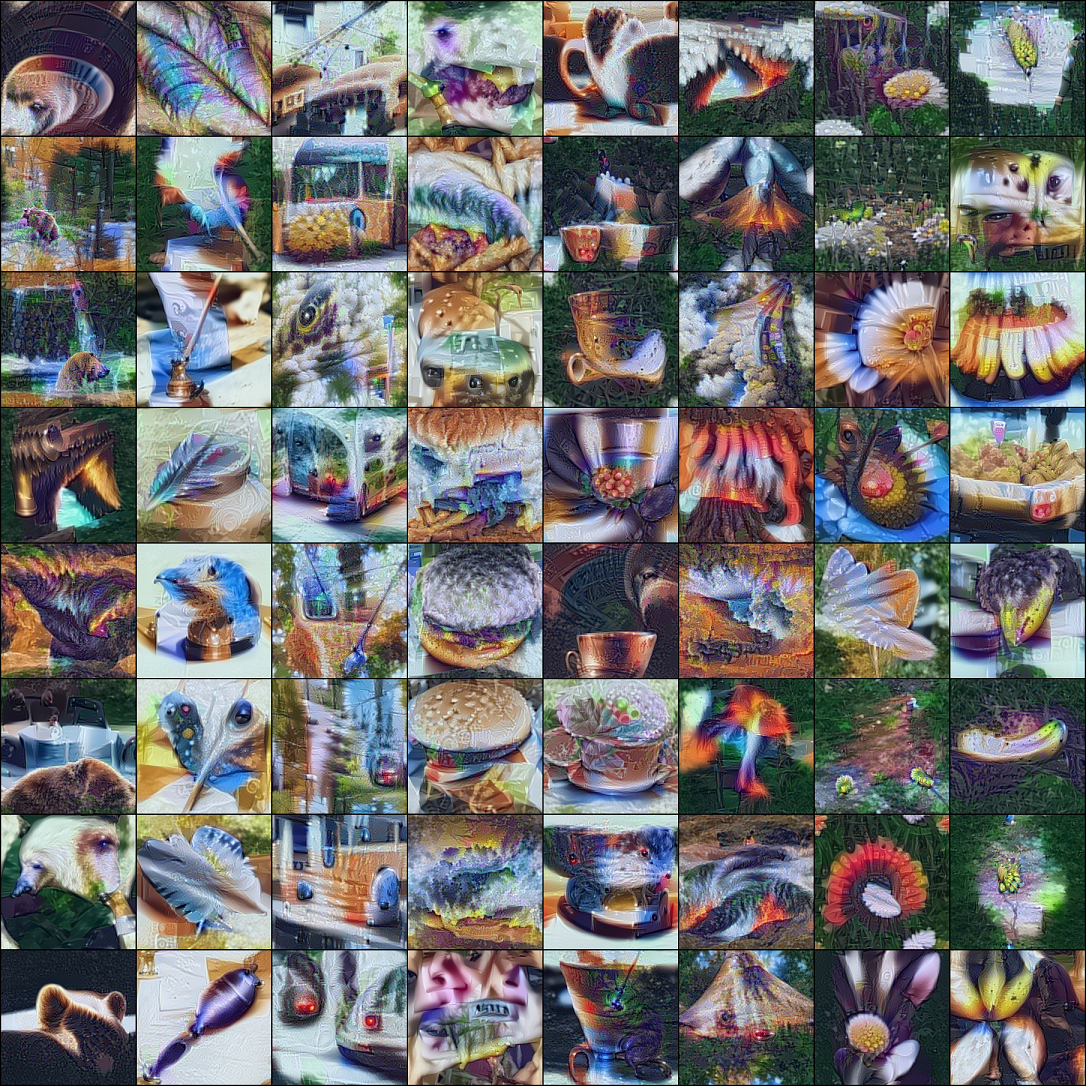}
    \end{minipage}
    }
    \caption{Class-conditional $224 \times 224$ images obtained by Deepinv with and without our adversarial loss given a resnet50 pre-trained on ImageNet. The classes left to right are brown bear, quill, trolleybus, cheeseburger, cup, volcano, daisy, banana. $\beta$ indicates the loss weight.} 
    \label{fig:imagenet-visualization} 
\end{figure}

As shown in Figure \ref{fig:l1-distance-imagenet}, we perform a quantitative analysis of how our main component $\mathcal{L}_{n t r i}$ affects the intra-class diversity and inter-class confusion of synthetic samples. In the embedding space, our method shows larger average distances among the embeddings of samples with the same labels, leading to higher intra-class diversity. Besides, our method shows smaller average distances among the embeddings of samples with different labels, leading to higher inter-class confusion. Meanwhile, we also analyze the sampling strategy of negative samples mentioned in Section \ref{sec:sampling}. We show how the negative sampling strategies in the image synthesis phase affect global diversity in Figure \ref{fig:global-distance-imagenet}. To be specific, synthesized samples with focal weighted sampling tend to have higher global diversity than those with distance weighted sampling. Because focal weighted sampling focuses on the negative only in local views, thus avoiding the potential damage to global diversity caused by pulling close the anchor and the negative already far away or near enough.

\noindent\textbf{Image Visualization}. In this section, we apply our method to a resnet50 trained on ImageNet to synthesize images for visualization. For fair comparison, we adopt two practical considerations from Deepinversion for high-resolution images, which are image clipping and multi-resolution synthesis. A set of images generated by Deepinv and our proposed RGAL on the pre-trained resnet50 is shown in Figure \ref{fig:imagenet-visualization}. Our RGAL synthesizes patterns and textures for more confusing samples, such as the daisy and banana on the right. The samples of the same classes are also more diverse. Even if the images by Deepinv demonstrate high fidelity and diversity, this is not necessary for knowledge transfer in the teacher-student model. We prefer the teacher to guide the student to understand the misclassified samples correctly. 

\begin{figure}
    \centering 
    \subfigure[cGAN]{
    \begin{minipage}{9cm}
    \centering
    \includegraphics[width=0.95\textwidth]{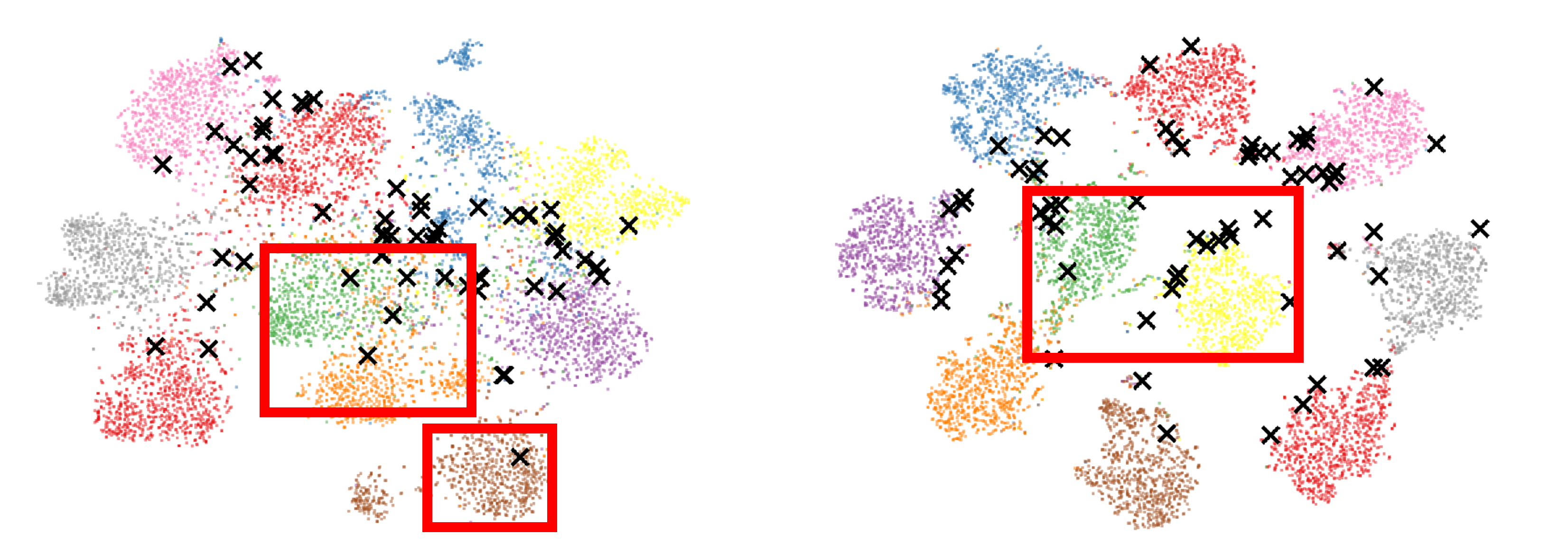}
    \end{minipage}
    }
    \subfigure[Deepinv]{
    \begin{minipage}{9cm}
    \centering  
    \includegraphics[width=0.95\textwidth]{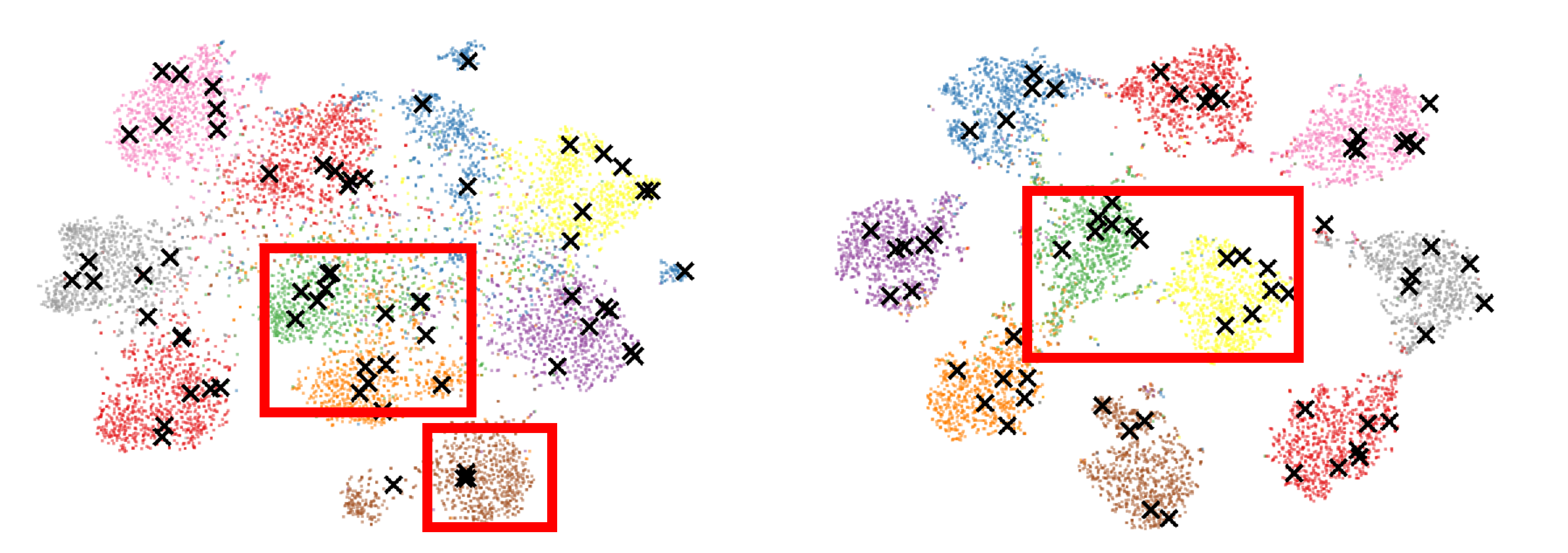}
    \end{minipage}
    }
    \subfigure[RGAL (ours)]{
    \begin{minipage}{9cm}
    \centering  
    \includegraphics[width=0.95\textwidth]{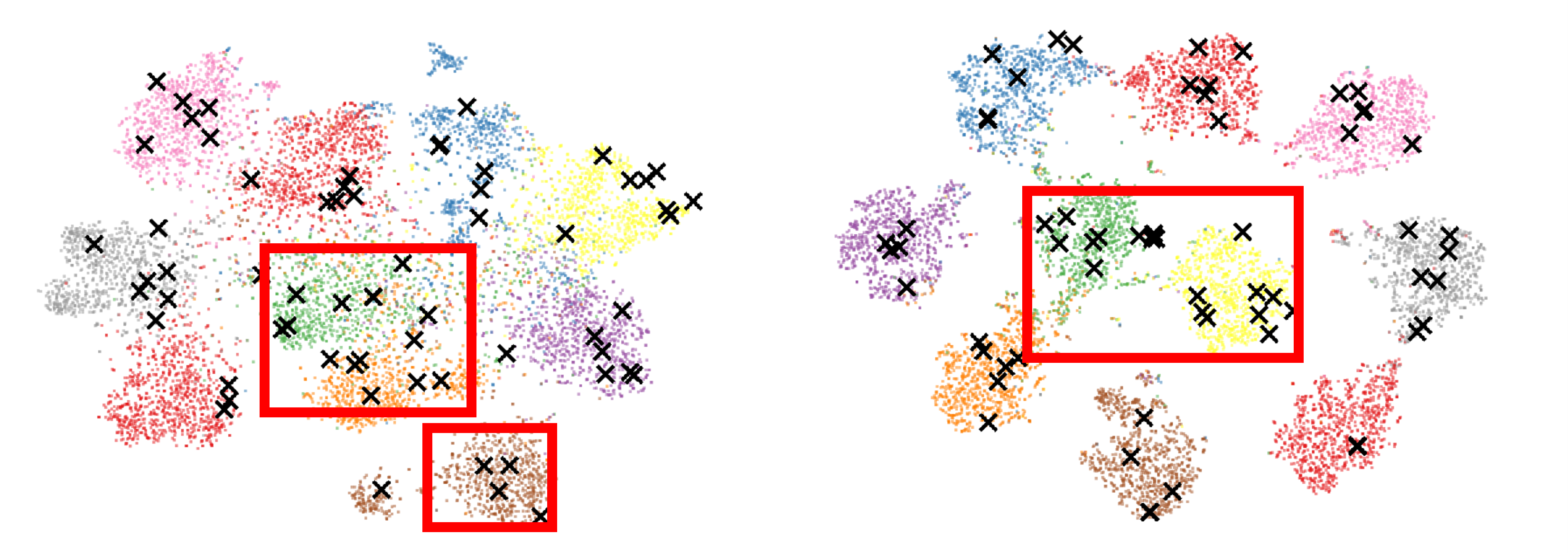}
    \end{minipage}
    }
    \caption{The distribution of global embedding from the intermediate layer (left) and the last layer (right). The background points are the real embedding distribution of the test set, and the black cross is the embedding of the generated images. The occurrence of embedding clusters is effectively reduced in RGAL when compared with other methods.}
    \label{fig:tsne-dis}
\end{figure}

\noindent\textbf{Distribution Visualization}. In this section, we apply our method to resnet34 trained on CIFAR-10 to synthesize 64 images. We compare the distribution of images generated by cGAN\citep{mirza2014conditional,chen2019data,xiang2019incremental} and Deepinv in the global embedding space and output embedding space. Both DAFL and ZSKT use cGAN to synthesize images, while the optimization tricks proposed by DeepInv are widely adopted by recent state-of-the-art methods, e.g., CMI and Fast-DFKD. T-SNE \citep{van2008visualizing} is used for embedding distribution visualization, as shown in Figure \ref{fig:tsne-dis}. 

Although cGAN generates samples with better visual quality, the distribution of these samples in the embedding space is too dense to cover the distribution space effectively. This can seriously affect the performance of the student  because the student can only learn from a limited part of the sample distribution. Deepinv extends the image regularization to solve the above problem. However, since the regularization does not constrain the relation among samples, embedding clusters still appear, decreasing sample diversity. On the other hand, RGAL maintains high intra-class diversity and improves inter-class confusion by relation-guided adversarial learning and inter-sample mutual constraints.

\section{Extensive Applications}

\subsection{Large-Scale Data-Free Distillation}

To further confirm whether the proposed method still works on large-scale datasets, we conduct data-free knowledge distillation on the full ImageNet. We follow the setup of Fast-DFKD \citep{fang2022up} and Deepinv \citep{bhardwaj2019dream} to implement our proposed RGAL. To be specific, considering that the number of classes ($1,000$) is much larger than the batch size ($100$), directly implementing RGAL on ImageNet without such data still leads to overfitting. Therefore, we generate the pre-synthesized data as in Fast-DFKD and Deepinv. We use a resnet-50 pre-trained on ImageNet to generate 140k samples following the official implementation of Deepinv to initialize the pre-synthesized data. Then we train the model using the core components proposed in our RGAL. We evaluate the model on ImageNet and show the results of RGAL with and without pre-synthesized data in Table \ref{tab:imagenet} and Table \ref{tab:imagenet2}, respectively. Specifically, the models trained with the ensemble of RGAL outperform the ones trained with the previous state-of-the-art methods, achieving an average accuracy improvement of $+0.59\%$ with pre-synthesized data. Besides, when not using pre-synthesized data, RGAL outperforms Fast-DFKD with an average accuracy improvement of +0.66\%. The results verify the effectiveness of our proposed RGAL on the large-scale dataset.

\begin{table}
  \centering
  \caption{Results of RGAL on large-scale full ImageNet dataset with pre-synthesized data. ``-" indicates no results reported in the paper.}
    \begin{tabular}{@{}lccc@{}}
    \midrule
    \multirow{2}{*}{Method}  & resnet50 & resnet50 & resnet50 \\
    ~ & resnet50 & resnet18 & mobilenetv2 \\ \midrule
    % Generative DFD & 69.75 & 54.66 & 43.15 \\
    DeepInv & 68.00 & - & - \\
    Fast-DFKD & 68.61 & 53.45 & 43.02 \\
    RGAL & \textbf{68.65} & \textbf{54.41} & \textbf{43.79} \\ \midrule 
  \end{tabular}
  \label{tab:imagenet}
\end{table}

\begin{table}
  \centering
  \caption{Results of RGAL on large-scale full ImageNet dataset without pre-synthesized data.}
    \begin{tabular}{@{}lccc@{}}
    \midrule
    \multirow{2}{*}{Method}  & resnet50 & resnet50 & resnet50 \\ 
    ~ & resnet50 & resnet18 & mobilenetv2 \\ \midrule
    Fast-DFKD & 67.16 & 42.98 & 38.05 \\
    RGAL & \textbf{67.38} & \textbf{43.64} & \textbf{39.15} \\ \midrule 
  \end{tabular}
  \label{tab:imagenet2}
\end{table}

\begin{figure*}[t]
    \centering
   \includegraphics[width=0.95\textwidth]{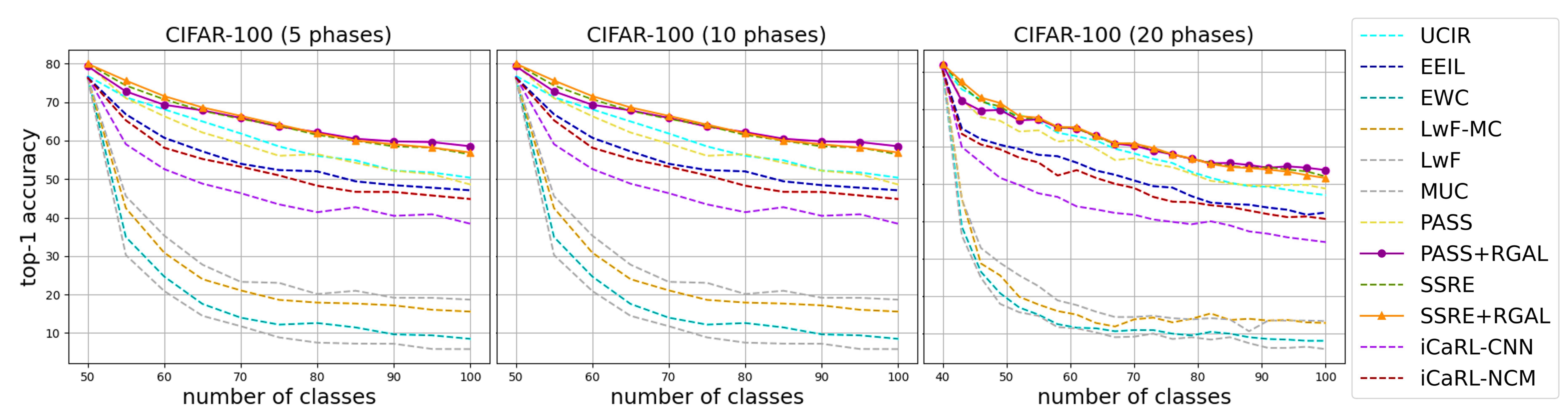}
    \caption{Classification accuracy results of 5, 10, and 20 incremental phases performed on CIFAR-100.} 
    \label{fig:cifar100-task-plot-visualization}  
\end{figure*}

\begin{figure*}[t]
    \centering
   \includegraphics[width=0.95\textwidth]{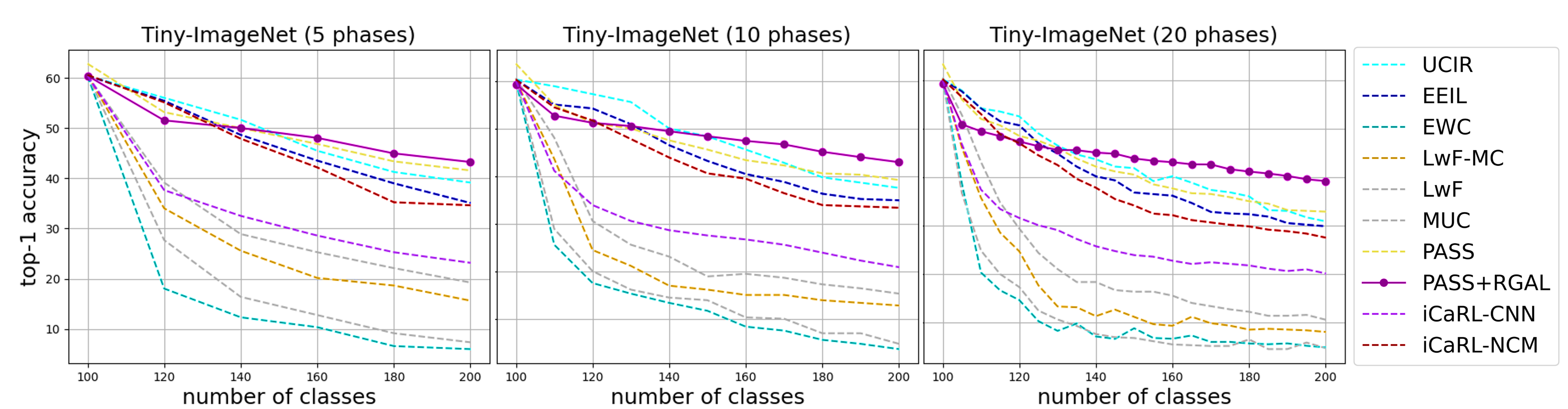}
    \caption{Classification accuracy results of 5, 10, and 20  sequential phases performed on Tiny-Imagenet.} 
    \label{fig:Tiny-ImageNet-task-plot-visualization} 
\end{figure*}

\subsection{Results on Data-Free Quantization}
\label{sec:dfq}
To further verify the effectiveness of our proposed RGALs, we conduct experiments on the data-free quantization task, which aims to quantize deep neural networks that do not require fine-tuning on real data. In this section, we formulate the data-free quantization process following the setup of IntraQ \citep{zhong2021intraq} and GDFQ \citep{xu2020generative}. IntraQ and GDFQ are the representative and state-of-the-art data-free quantization methods with and without image generators, respectively. The teacher $\mathcal{T}$ is initially well-trained on the entire real training set and the student $\mathcal{S}$ is the quantized model with low-precision fixed-point operations and is trained with syntactic data only.

To perform our proposed RGAL on these two methods, the generator or image synthesis optimizer is trained with an additional negative relation adversarial loss:
\begin{equation}
  \begin{aligned}
  \mathcal{L}^{\prime}_{s} = - \mathcal{L}_{a d v} + \mathcal{L}_{n t r i},
  \label{eq:training-generator-quant-neg}
  \end{aligned}
\end{equation}
while the student is trained with:
\begin{equation}
  \begin{aligned}
  \mathcal{L}^{\prime}_{g} = \mathcal{L}_{a d v} + \mathcal{L}_{t r i} + \mathcal{L}_{o h}.
  \label{eq:training-generator-quant-pos}
  \end{aligned}
\end{equation}

\begin{table}
  \centering
  \caption{Quantization results of students without and with our formulation. The teacher is a resnet-20 trained on CIFAR-100. The weights and activations are quantized to 4-bit and 3-bit.}
    \begin{tabular}{@{}lccc@{}}
    \midrule
    Bit-width & Method & Generator & Top1-Acc \\ \midrule
    \multirow{4}{*}{W4A4} & IntraQ & $\times$ & 64.98 \\ 
    ~ & \cellcolor{gray!20}IntraQ+RGAL & \cellcolor{gray!20}$\times$ & \cellcolor{gray!20}\textbf{65.10 (+0.12)} \\
    ~ & GDFQ & $\checkmark$ & 63.58 \\ 
    ~ & \cellcolor{gray!20}GDFQ+RGAL & \cellcolor{gray!20}$\checkmark$ & \cellcolor{gray!20}\textbf{63.95 (+0.37)} \\ \midrule
    \multirow{4}{*}{W3A3} & IntraQ & $\times$ & 48.25 \\ 
    ~ & \cellcolor{gray!20}IntraQ+RGAL & \cellcolor{gray!20}$\times$ & \cellcolor{gray!20}\textbf{49.34 (+1.09)} \\ 
    ~ & GDFQ & $\checkmark$ & 43.87 \\ 
    ~ & \cellcolor{gray!20}GDFQ+RGAL & \cellcolor{gray!20}$\checkmark$ &\cellcolor{gray!20}\textbf{47.53 (+3.66)} \\ \midrule
  \end{tabular}
  \label{tab:quant-cifar100}
\end{table}

\begin{table}
  \centering
  \caption{Quantization results of IntraQ with and without RGAL on ImageNet. The teacher before quantization is a resnet-18 with 76.13$\%$ top-1 accuracy.}
    \begin{tabular}{@{}lccc@{}}
    \midrule
    Bit-width & $\mathcal{L}^{\prime}_{g}$ & $\mathcal{L}^{\prime}_{s}$ & Top1-Acc \\ \midrule
    \multirow{3}{*}{W5A5} & $\times$ & $\times$ & 69.94 \\ 
    ~ & $\checkmark$ & $\times$ & 70.78 \\ 
    ~ & $\checkmark$ & $\checkmark$ & \textbf{70.85} \\ \midrule
    \multirow{3}{*}{W4A4} & $\times$ & $\times$ & 66.47 \\ 
    ~ & $\checkmark$ & $\times$ & 66.84 \\ 
    ~ & $\checkmark$ & $\checkmark$ & \textbf{66.85} \\ \midrule
  \end{tabular}
  \label{tab:imagenet-results}
\end{table}

To quantify the benefits of our proposed RGAL for these methods, we first explore the data-free network quantization task under 4-bit and 3-bit settings on the CIFAR-100 dataset. We show the accuracy of quantized resnet-20 trained on prior works with and without RGAL. Table \ref{tab:quant-cifar100} shows the accuracy of quantization after adding RGAL to IntraQ and GDFQ. IntraQ and GDFQ with RGAL achieve accuracy improvements of $+0.12\%$ and $+0.37\%$ under 4-bit settings, respectively, and gets accuracy improvements of $+1.09\%$ and $+3.66\%$ under more difficult 3-bit settings.

To confirm whether our proposed method still works on large-scale datasets, we conduct data-free quantization on the full ImageNet based on IntraQ. Similar to the experiment on CIFAR-100, we incrementally add to IntraQ the losses we are using in the RGAL. Table \ref{tab:imagenet-results} shows the accuracy of resnet-18 trained on IntraQ with and without the proposed losses, i.e., $\mathcal{L}^{\prime}_{g}$ for image synthesizing and $\mathcal{L}^{\prime}_{s}$ for training the student (quantized) model. Specifically, our proposed RGAL improves IntraQ with resnet-18 on the ImageNet dataset with accuracy gains of +0.91\% and +0.38\% under 5-bit and 4-bit settings, respectively.

\begin{table*}
  \caption{Average accuracy of resnet-18 on Tiny-Imagenet and CIFAR-100. For incremental learning, ``Phases'' indicates the number of tasks.}
  \centering
  \begin{tabular}{@{}lcccccc@{}}
    \midrule
    \multirow{2}{*}{Method} & \multicolumn{3}{c}{CIFAR-100} & \multicolumn{3}{c}{Tiny-ImageNet} \\ \cmidrule(lr){2-4}  \cmidrule(lr){5-7} 
    ~ & 5 phases & 10 phases & 20 phases & 5 phases & 10 phases & 20 phases \\ \midrule
    LwF-MC & 45.93 & 27.43 & 20.07 & 29.12 & 23.10 & 17.43 \\ 
    MUC & 49.42 & 30.19 & 21.27 & 32.58 & 26.61 & 21.95 \\ 
    iCaRL-CNN & 51.07 & 48.66 & 44.43 & 34.64 & 31.15 & 27.90 \\ 
    iCaRL-NCM & 58.56 & 54.19 & 50.51 & 45.86 & 43.29 & 38.04 \\
    EEIL & 60.37 & 56.05 & 52.34 & 47.12 & 45.01 & 40.50 \\ 
    UCIR & 63.78 & 62.39 & 59.07 & 49.15 & 48.52 & 42.83 \\ 
    PASS & 63.47 & 61.84 & 58.09 & 49.55 & 47.29 & 42.07 \\
    \cellcolor{gray!20}PASS+RGAL & \cellcolor{gray!20}\textbf{66.95 (+3.48)} & \cellcolor{gray!20}\textbf{65.47 (+3.63)} & \cellcolor{gray!20}\textbf{61.88 (+3.79)} & \cellcolor{gray!20}\textbf{49.82 (+0.27)} & \cellcolor{gray!20}\textbf{48.86 (+1.57)} & \cellcolor{gray!20}\textbf{44.80 (+2.73)} \\ 
    SSRE & 65.88 & 65.04 & 61.70 & N/A & N/A & N/A \\ 
    \cellcolor{gray!20}SSRE+RGAL & \cellcolor{gray!20}\textbf{66.42 (+0.54)} & \cellcolor{gray!20}\textbf{65.68 (+0.64)} & \cellcolor{gray!20}\textbf{62.17 (+0.47)} & N/A & N/A & N/A \\ \midrule
  \end{tabular}
  \label{tab:average-acc-results}
\end{table*}

\begin{table*}
  \caption{Average forgetting of resnet-18 on Tiny-Imagenet and CIFAR-100. A lower average forgetting indicates better performance. The results show that our method can effectively prevent information forgetting.}
  \centering
  \begin{tabular}{@{}lcccccc@{}}
    \midrule
    \multirow{2}{*}{Method} & \multicolumn{3}{c}{CIFAR-100} & \multicolumn{3}{c}{Tiny-ImageNet} \\ \cmidrule(lr){2-4}  \cmidrule(lr){5-7} 
    ~ & 5 phases & 10 phases & 20 phases & 5 phases & 10 phases & 20 phases \\ \midrule
    LwF-MC & 44.23 & 50.47 & 55.46 & 54.26 & 54.37 & 63.54 \\ \midrule
    MUC & 40.28 & 47.56 & 52.63 & 51.46 & 50.21 & 58.00 \\ \midrule
    iCaRL-CNN & 42.13 & 45.69 & 43.54 & 36.89 & 36.70 & 45.12 \\ \midrule
    iCaRL-NCM & 24.90 & 28.32 & 35.53 & 27.15 & 28.89 & 37.40 \\ \midrule
    EEIL & 23.36 & 26.65 & 32.40 & 25.56 & 25.91 & 35.04 \\ \midrule
    UCIR & 21.00 & 25.12 & 28.65 & 20.61 & 22.25 & 33.74 \\ \midrule
    PASS & 25.20 & 30.25& 30.61 & 18.04 & 23.11 & 30.55 \\ 
   \cellcolor{gray!20} PASS+RGAL & \cellcolor{gray!20}\textbf{17.19 (-8.01)} & \cellcolor{gray!20}\textbf{22.16 (-8.09)} & \cellcolor{gray!20}\textbf{24.27 (-6.34)} & \cellcolor{gray!20}\textbf{15.36 (-2.68)} & \cellcolor{gray!20}\textbf{20.28 (-2.83)} & \cellcolor{gray!20}\textbf{23.63 (-6.92)} \\ \midrule
    SSRE & 18.37 &  19.48 & 19.00 & N/A & N/A & N/A \\ 
    \cellcolor{gray!20}SSRE+RGAL & \cellcolor{gray!20}\textbf{12.17 (-6.20)} & \cellcolor{gray!20}\textbf{10.64 (-8.84)} & \cellcolor{gray!20}\textbf{7.15 (-11.85)} & N/A & N/A & N/A \\ \midrule
  \end{tabular}
  \label{tab:forgetting-results}
\end{table*}

\subsection{Results on Non-Exemplar Incremental Learning}
\label{sec:dfil}

Incremental learning allows for efficient resource usage by eliminating the need to retrain from scratch at the arrival of new data and reduced memory usage by limiting the amount of previous data required to be stored. However, in most cases, small amounts of previous data are also unavailable due to storage limitations or data privacy. To this end, non-exemplar incremental learning is proposed and aims to implement incremental learning without the need to retrain the model from scratch at the arrival of new data and by storing no previous data \citep{masana2020class}. We formulate the incremental learning process following the setup of PASS \citep{zhu2021prototype} and SSRE \citep{zhu2022self}. Specifically, PASS and SSRE train the model with data of new classes and a small number of saved prototypes from old data, where prototypes are equivalent to the global embeddings in our paper and are usually used as input to the classification head. In our setup, prototypes of old classes are optimized by the losses in Equation \eqref{eq:training-generator-quant-pos} for $10$ steps with a learning rate of 0.001 before training. Meanwhile, in incremental training, we also add an extra positive triplet loss $\mathcal{L}_{t r i}$ to train the model and help cluster embedding. This does not result in additional memory usage because we are not adding additional stored data. All other settings are the same as iCaRL. 

We consider state-of-the-art incremental learning methods such as EWC \citep{kirkpatrick2017overcoming}, LwF \citep{li2017learning}, LwF-MC \citep{rebuffi2017icarl}, LwM\citep{dhar2019learning}, MUC \citep{liu2020more}, PASS\citep{zhu2021prototype} and SSRE \citep{zhu2022self}. Moreover, we show the average accuracy of PASS and SSRE with and without our proposed RGAL in Table \ref{tab:average-acc-results}, which are state-of-the-art non-exemplar methods. Our proposed RGAL significantly improves the performance of models trained on PASS, with average accuracy improvements of +3.63\% on CIFAR-100 and +1.52\% on Tiny-ImageNet. Besides, RGAL also improves the performance of state-of-the-art SSRE with an average accuracy improvement of +0.55\% on CIFAR-100.

We also compare several of the most advanced exemplar-based methods, e.g., iCaRL \citep{rebuffi2017icarl}, EEIL \citep{castro2018end}, and UCIR \citep{hou2019learning}, as shown in Figures \ref{fig:cifar100-task-plot-visualization} and \ref{fig:Tiny-ImageNet-task-plot-visualization}. Methods with RGAL are superior at almost all phases, even compared with the methods requiring previous data. It can also be seen that the benefit of our formulation gradually increases as the number of incremental phases increases with more new classes. 

\begin{table}
  \centering
  \caption{Overall accuracy of resnet-18 on PASS without and with the core components in RGAL.}
  \begin{tabular}{@{}lccccc@{}}
    \midrule
    \multirow{2}{*}{Dataset} & \multicolumn{2}{c}{Losses} & \multicolumn{3}{c}{Phases} \\  \cmidrule(lr){2-3}  \cmidrule(lr){4-6} 
    ~ & $\mathcal{L}^{\prime}_{g}$ & $\mathcal{L}^{\prime}_{s}$ & 5 & 10 & 20 \\ \midrule
    \multirow{3}{*}{CIFAR-100} & × & × & 55.67 & 49.03 & 48.48 \\ 
    ~ & $\checkmark$ & × & 59.18 & 57.69 & 52.90 \\ 
    ~ & $\checkmark$ & $\checkmark$ & \textbf{59.70} & \textbf{58.54} & \textbf{53.75} \\ \midrule
    \multirow{3}{*}{Tiny-ImageNet} & × & × & 41.58 & 39.28 & 32.78 \\ 
    ~ & $\checkmark$ & × & 43.09 & 42.01 & \textbf{41.03} \\  
    ~ & $\checkmark$ & $\checkmark$ & \textbf{43.27} & \textbf{42.91} & 39.19 \\ \midrule
  \end{tabular}
  \label{tab:incremental-results}
\end{table}

\begin{table}
  \centering
  \caption{Average accuracy of resnet-18 on PASS without and with the core components in RGAL.}
  \begin{tabular}{@{}lccccc@{}}
    \midrule
    \multirow{2}{*}{Dataset} & \multicolumn{2}{c}{Losses} & \multicolumn{3}{c}{Phases} \\ \cmidrule(lr){2-3}  \cmidrule(lr){4-6} 
    ~ & $\mathcal{L}^{\prime}_{g}$ & $\mathcal{L}^{\prime}_{s}$ & 5 & 10 & 20 \\ \midrule 
    \multirow{3}{*}{CIFAR-100} & × & × & 63.47 & 61.84 & 58.09 \\ 
    ~ & $\checkmark$ & × & 66.75 & 65.08 & 61.60 \\ 
    ~ & $\checkmark$ & $\checkmark$ & \textbf{66.95} & \textbf{65.47} & \textbf{61.88} \\ \midrule
    \multirow{3}{*}{Tiny-ImageNet} & × & × & 49.55 & 47.29 & 42.07 \\
    ~ & $\checkmark$ & × & 49.68 & 47.74 & \textbf{47.37} \\ 
    ~ & $\checkmark$ & $\checkmark$ & \textbf{49.82} & \textbf{48.86} & 44.80 \\ \midrule
  \end{tabular}
  \label{tab:incremental-average-accuracy}
\end{table}

Average forgetting \citep{chaudhry2018riemannian} shows how the model forgets the previous phase and a lower average forgetting indicates the model forgets less about the old knowledge. To verify the effectiveness of our proposed RGAL from another perspective, we also show the average forgetting results of PASS and SSRE with and without our proposed RGAL in Table \ref{tab:forgetting-results}. Pass with RGAL achieves average forgetting improvements of $-7.44\%$ on CIFAR-100 and $-4.14\%$ on Tiny-ImageNet. Besides, SSRE with RGAL achieves an average forgetting improvement of $-8.96\%$ on CIFAR-100. The results show that methods with RGAL can significantly prevent information forgetting and adapt to new phases more effectively with higher accuracy. 

Tables \ref{tab:incremental-results} and \ref{tab:incremental-average-accuracy} also report the overall accuracy and average accuracy of all the classes already learned on PASS without and with our proposed two adversarial losses. The former represents the accuracy of the model in all classes after training in the last phase, while the latter represents the average accuracy of all phases. As the number of tasks increases, the benefit of losses proposed in RGAL increases, except $\mathcal{L}_{s}^{\prime}$ on Tiny-ImageNet. The results show that both losses proposed in our RGAL, $\mathcal{L}_{g}^{\prime}$ and $\mathcal{L}_{s}^{\prime}$, have beneficial effects on incremental learning phases. 

\section{Conclusion}

In this paper, we design a novel relation-guided adversarial learning method, namely RGAL, for data-free knowledge transfer. Our proposed RGAL locates and addresses the key challenge by seeking intra-class diversity and inter-class confusion of samples within a batch at the instance level. We also present a focal weighted strategy to mitigate the potential optimization conflicts between global diversity and inter-class confusion. Extensive experiments and analysis on data-free knowledge distillation demonstrate the effectiveness of our method. Furthermore, the application of RGAL to other tasks, specifically data-free quantization and non-exemplar incremental learning, indicates its robust generalizability and significant enhancements in data-free knowledge transfer applications.

\noindent\textbf{Limitations.} Our study encounters a common limitation for data-free knowledge distillation wherein the student model does not attain the accuracy level of models trained on real datasets.  Additionally, our focus is predominantly on classification tasks, and we have not ventured into areas such as object detection, semantic segmentation, etc. Moving forward, we aim to delve deeper into understanding the disparity between data-free knowledge distillation and training with real data, and intend to broaden the applicability of our method to encompass a wider range of tasks.

\noindent\textbf{Data Availability.} Data sharing does not apply to this article, as no datasets were generated or analyzed during the current study.

\noindent\textbf{Acknowledgements.} This work was supported by the National Natural Science Foundation of China (62331006, 6217 1038, 61931008, and 62088101), and the Fundamental Research Funds for the Central Universities.

\bibliographystyle{spbasic} 
\bibliography{sample}{}

\begin{thebibliography}{59}
\providecommand{\natexlab}[1]{#1}
\providecommand{\url}[1]{{#1}}
\providecommand{\urlprefix}{URL }
\expandafter\ifx\csname urlstyle\endcsname\relax
  \providecommand{\doi}[1]{DOI~\discretionary{}{}{}#1}\else
  \providecommand{\doi}{DOI~\discretionary{}{}{}\begingroup \urlstyle{rm}\Url}\fi
\providecommand{\eprint}[2][]{\url{#2}}

\bibitem[{Arora and Bhatia(2021)}]{arora2020biometric}
Arora S, Bhatia M (2021) A secure framework to preserve privacy of biometric templates on cloud using deep learning. Recent Advances in Computer Science and Communications 14(5):1412--1421

\bibitem[{Ba and Caruana(2014)}]{ba2013deep}
Ba J, Caruana R (2014) Do deep nets really need to be deep? In: Advances in Neural Information Processing Systems, pp 2654--2662

\bibitem[{Bhardwaj et~al(2019)Bhardwaj, Suda, and Marculescu}]{bhardwaj2019dream}
Bhardwaj K, Suda N, Marculescu R (2019) Dream distillation: A data-independent model compression framework. In: International Conference on Machine Learning Workshop

\bibitem[{Bucilua et~al(2006)Bucilua, Caruana, and Niculescu-Mizil}]{bucilua2006model}
Bucilua C, Caruana R, Niculescu-Mizil A (2006) Model compression. In: International Conference on Knowledge Discovery and Data Mining, pp 535--541

\bibitem[{Cai et~al(2020)Cai, Yao, Dong, Gholami, Mahoney, and Keutzer}]{cai2020zeroq}
Cai Y, Yao Z, Dong Z, Gholami A, Mahoney MW, Keutzer K (2020) Zeroq: A novel zero shot quantization framework. In: the IEEE Conference on Computer Vision and Pattern Recognition, pp 13,169--13,178

\bibitem[{Castro et~al(2018)Castro, Mar{\'\i}n-Jim{\'e}nez, Guil, Schmid, and Alahari}]{castro2018end}
Castro FM, Mar{\'\i}n-Jim{\'e}nez MJ, Guil N, Schmid C, Alahari K (2018) End-to-end incremental learning. In: the European Conference on Computer Vision, pp 233--248

\bibitem[{Chaudhry et~al(2018)Chaudhry, Dokania, Ajanthan, and Torr}]{chaudhry2018riemannian}
Chaudhry A, Dokania PK, Ajanthan T, Torr PH (2018) Riemannian walk for incremental learning: Understanding forgetting and intransigence. In: the European Conference on Computer Vision, pp 532--547

\bibitem[{Chen et~al(2019)Chen, Wang, Xu, Yang, Liu, Shi, Xu, Xu, and Tian}]{chen2019data}
Chen H, Wang Y, Xu C, Yang Z, Liu C, Shi B, Xu C, Xu C, Tian Q (2019) Data-free learning of student networks. In: the IEEE International Conference on Computer Vision, pp 3514--3522

\bibitem[{Cheng et~al(2018)Cheng, Wang, Zhou, and Zhang}]{cheng2018model}
Cheng Y, Wang D, Zhou P, Zhang T (2018) Model compression and acceleration for deep neural networks: principles, progress, and challenges. IEEE Signal Processing Magazine 35(1):126--136

\bibitem[{Choi et~al(2020)Choi, Choi, El-Khamy, and Lee}]{choi2020data}
Choi Y, Choi J, El-Khamy M, Lee J (2020) Data-free network quantization with adversarial knowledge distillation. In: the IEEE Conference on Computer Vision and Pattern Recognition Workshops, pp 710--711

\bibitem[{Deng et~al(2009)Deng, Dong, Socher, Li, Li, and Fei-Fei}]{deng2009imagenet}
Deng J, Dong W, Socher R, Li LJ, Li K, Fei-Fei L (2009) Imagenet: A large-scale hierarchical image database. In: the IEEE Conference on Computer Vision and Pattern Recognition, pp 248--255

\bibitem[{Dhar et~al(2019)Dhar, Singh, Peng, Wu, and Chellappa}]{dhar2019learning}
Dhar P, Singh RV, Peng KC, Wu Z, Chellappa R (2019) Learning without memorizing. In: the IEEE Conference on Computer Vision and Pattern Recognition, pp 5138--5146

\bibitem[{Fang et~al(2019)Fang, Song, Shen, Wang, Chen, and Song}]{fang2019data}
Fang G, Song J, Shen C, Wang X, Chen D, Song M (2019) Data-free adversarial distillation. arXiv preprint arXiv:191211006

\bibitem[{Fang et~al(2021)Fang, Song, Wang, Shen, Wang, and Song}]{fang2021contrastive}
Fang G, Song J, Wang X, Shen C, Wang X, Song M (2021) Contrastive model inversion for data-free knowledge distillation. In: International Joint Conference on Artificial Intelligence, pp 2374--2380

\bibitem[{Fang et~al(2022)Fang, Mo, Wang, Song, Bei, Zhang, and Song}]{fang2022up}
Fang G, Mo K, Wang X, Song J, Bei S, Zhang H, Song M (2022) Up to 100x faster data-free knowledge distillation. In: the AAAI Conference on Artificial Intelligence, vol~36, pp 6597--6604

\bibitem[{Girshick et~al(2014)Girshick, Donahue, Darrell, and Malik}]{girshick2014rich}
Girshick R, Donahue J, Darrell T, Malik J (2014) Rich feature hierarchies for accurate object detection and semantic segmentation. In: the IEEE Conference on Computer Vision and Pattern Recognition, pp 580--587

\bibitem[{Goodfellow et~al(2014)Goodfellow, Pouget-Abadie, Mirza, Xu, Warde-Farley, Ozair, Courville, and Bengio}]{goodfellow2014generative}
Goodfellow I, Pouget-Abadie J, Mirza M, Xu B, Warde-Farley D, Ozair S, Courville A, Bengio Y (2014) Generative adversarial nets. In: Advances in Neural Information Processing Systems, pp 2672--2680

\bibitem[{Gou et~al(2021)Gou, Yu, Maybank, and Tao}]{gou2021knowledge}
Gou J, Yu B, Maybank SJ, Tao D (2021) Knowledge distillation: A survey. International Journal of Computer Vision 129(6):1789--1819

\bibitem[{Ha et~al(2020)Ha, Dang, Le, and Truong}]{ha2020security}
Ha T, Dang TK, Le H, Truong TA (2020) Security and privacy issues in deep learning: a brief review. SN Computer Science 1(5):1--15

\bibitem[{Han et~al(2021)Han, Park, Wang, and Liu}]{han2021robustness}
Han P, Park J, Wang S, Liu Y (2021) Robustness and diversity seeking data-free knowledge distillation. In: the IEEE International Conference on Acoustics, Speech and Signal Processing, pp 2740--2744

\bibitem[{He et~al(2016)He, Zhang, Ren, and Sun}]{he2016deep}
He K, Zhang X, Ren S, Sun J (2016) Deep residual learning for image recognition. In: the IEEE Conference on Computer Vision and Pattern Recognition, pp 770--778

\bibitem[{Hinton et~al(2014)Hinton, Vinyals, and Dean}]{hinton2015distilling}
Hinton G, Vinyals O, Dean J (2014) Distilling the knowledge in a neural network. In: Deep Learning and Representation Learning Workshop

\bibitem[{Hou et~al(2019)Hou, Pan, Loy, Wang, and Lin}]{hou2019learning}
Hou S, Pan X, Loy CC, Wang Z, Lin D (2019) Learning a unified classifier incrementally via rebalancing. In: the IEEE Conference on Computer Vision and Pattern Recognition, pp 831--839

\bibitem[{Kingma and Ba(2015)}]{kingma2014adam}
Kingma DP, Ba J (2015) Adam: A method for stochastic optimization. In: International Conference on Learning Representations

\bibitem[{Kirkpatrick et~al(2017)Kirkpatrick, Pascanu, Rabinowitz, Veness, Desjardins, Rusu, Milan, Quan, Ramalho, Grabska-Barwinska et~al}]{kirkpatrick2017overcoming}
Kirkpatrick J, Pascanu R, Rabinowitz N, Veness J, Desjardins G, Rusu AA, Milan K, Quan J, Ramalho T, Grabska-Barwinska A, et~al (2017) Overcoming catastrophic forgetting in neural networks. The National Academy of Sciences 114(13):3521--3526

\bibitem[{Krizhevsky and Hinton(2009)}]{krizhevsky2009learning}
Krizhevsky A, Hinton G (2009) Learning multiple layers of features from tiny images. Handbook of Systemic Autoimmune Diseases 1(4)

\bibitem[{Krizhevsky et~al(2012)Krizhevsky, Sutskever, and Hinton}]{krizhevsky2012imagenet}
Krizhevsky A, Sutskever I, Hinton GE (2012) Imagenet classification with deep convolutional neural networks. In: Advances in Neural Information Processing Systems, pp 1097--1105

\bibitem[{Li and Hoiem(2017)}]{li2017learning}
Li Z, Hoiem D (2017) Learning without forgetting. IEEE Transactions on Pattern Analysis and Machine Intelligence 40(12):2935--2947

\bibitem[{Liu et~al(2020)Liu, Parisot, Slabaugh, Jia, Leonardis, and Tuytelaars}]{liu2020more}
Liu Y, Parisot S, Slabaugh G, Jia X, Leonardis A, Tuytelaars T (2020) More classifiers, less forgetting: A generic multi-classifier paradigm for incremental learning. In: the European Conference on Computer Vision, pp 699--716

\bibitem[{Liu et~al(2021)Liu, Zhang, and Wang}]{liu2021zero}
Liu Y, Zhang W, Wang J (2021) Zero-shot adversarial quantization. In: the IEEE Conference on Computer Vision and Pattern Recognition, pp 1512--1521

\bibitem[{Long et~al(2015)Long, Shelhamer, and Darrell}]{long2015fully}
Long J, Shelhamer E, Darrell T (2015) Fully convolutional networks for semantic segmentation. In: the IEEE Conference on Computer Vision and Pattern Recognition, pp 3431--3440

\bibitem[{Lopes et~al(2017)Lopes, Fenu, and Starner}]{lopes2017data}
Lopes RG, Fenu S, Starner T (2017) Data-free knowledge distillation for deep neural networks. In: Advances in Neural Information Processing Systems Workshop

\bibitem[{Loshchilov and Hutter(2017)}]{loshchilov2016sgdr}
Loshchilov I, Hutter F (2017) Sgdr: Stochastic gradient descent with warm restarts. In: International Conference on Learning Representations

\bibitem[{Van~der Maaten and Hinton(2008)}]{van2008visualizing}
Van~der Maaten L, Hinton G (2008) Visualizing data using t-sne. Journal of Machine Learning Research 9(11)

\bibitem[{Mahendran and Vedaldi(2015)}]{mahendran2015understanding}
Mahendran A, Vedaldi A (2015) Understanding deep image representations by inverting them. In: the IEEE Conference on Computer Vision and Pattern Recognition, pp 5188--5196

\bibitem[{Masana et~al(2020)Masana, Liu, Twardowski, Menta, Bagdanov, and van~de Weijer}]{masana2020class}
Masana M, Liu X, Twardowski B, Menta M, Bagdanov AD, van~de Weijer J (2020) Class-incremental learning: survey and performance evaluation on image classification. arXiv preprint arXiv:201015277

\bibitem[{Micaelli and Storkey(2019)}]{Micaelli2019ZeroShotKT}
Micaelli P, Storkey AJ (2019) Zero-shot knowledge transfer via adversarial belief matching. In: Advances in Neural Information Processing Systems, pp 9551--9561

\bibitem[{Mirza and Osindero(2014)}]{mirza2014conditional}
Mirza M, Osindero S (2014) Conditional generative adversarial nets. arXiv preprint arXiv:14111784

\bibitem[{Nayak et~al(2019)Nayak, Mopuri, Shaj, Radhakrishnan, and Chakraborty}]{nayak2019zero}
Nayak GK, Mopuri KR, Shaj V, Radhakrishnan VB, Chakraborty A (2019) Zero-shot knowledge distillation in deep networks. In: International Conference on Machine Learning, pp 4743--4751

\bibitem[{Nie and Shen(2020)}]{nie2020adversarial}
Nie D, Shen D (2020) Adversarial confidence learning for medical image segmentation and synthesis. International Journal of Computer Vision 128(10):2494--2513

\bibitem[{Park et~al(2019)Park, Kim, Lu, and Cho}]{park2019relational}
Park W, Kim D, Lu Y, Cho M (2019) Relational knowledge distillation. In: the IEEE Conference on Computer Vision and Pattern Recognition, pp 3967--3976

\bibitem[{Rebuffi et~al(2017)Rebuffi, Kolesnikov, Sperl, and Lampert}]{rebuffi2017icarl}
Rebuffi SA, Kolesnikov A, Sperl G, Lampert CH (2017) icarl: Incremental classifier and representation learning. In: the IEEE Conference on Computer Vision and Pattern Recognition, pp 2001--2010

\bibitem[{Redmon et~al(2016)Redmon, Divvala, Girshick, and Farhadi}]{redmon2016you}
Redmon J, Divvala S, Girshick R, Farhadi A (2016) You only look once: Unified, real-time object detection. In: the IEEE Conference on Computer Vision and Pattern Recognition, pp 779--788

\bibitem[{Romero et~al(2015)Romero, Ballas, Kahou, Chassang, Gatta, and Bengio}]{romero2014fitnets}
Romero A, Ballas N, Kahou SE, Chassang A, Gatta C, Bengio Y (2015) Fitnets: Hints for thin deep nets. In: International Conference on Learning Representations

\bibitem[{Ronneberger et~al(2015)Ronneberger, Fischer, and Brox}]{ronneberger2015u}
Ronneberger O, Fischer P, Brox T (2015) U-net: Convolutional networks for biomedical image segmentation. In: International Conference on Medical Image Computing and Computer-assisted Intervention, pp 234--241

\bibitem[{Schroff et~al(2015)Schroff, Kalenichenko, and Philbin}]{schroff2015facenet}
Schroff F, Kalenichenko D, Philbin J (2015) Facenet: A unified embedding for face recognition and clustering. In: the IEEE Conference on Computer Vision and Pattern Recognition, pp 815--823

\bibitem[{Simonyan and Zisserman(2015)}]{simonyan2014very}
Simonyan K, Zisserman A (2015) Very deep convolutional networks for large-scale image recognition. In: International Conference on Learning Representations

\bibitem[{Wu et~al(2017{\natexlab{a}})Wu, Manmatha, Smola, and Krahenbuhl}]{wu2017sampling}
Wu CY, Manmatha R, Smola AJ, Krahenbuhl P (2017{\natexlab{a}}) Sampling matters in deep embedding learning. In: the IEEE International Conference on Computer Vision, pp 2840--2848

\bibitem[{Wu et~al(2017{\natexlab{b}})Wu, Zhang, and Xu}]{wu2017tiny}
Wu J, Zhang Q, Xu G (2017{\natexlab{b}}) Tiny imagenet challenge. Technical Report

\bibitem[{Wu et~al(2020)Wu, He, Hu, and Sun}]{wu2020learning}
Wu X, He R, Hu Y, Sun Z (2020) Learning an evolutionary embedding via massive knowledge distillation. International Journal of Computer Vision 128(8):2089--2106

\bibitem[{Xiang et~al(2019)Xiang, Fu, Ji, and Huang}]{xiang2019incremental}
Xiang Y, Fu Y, Ji P, Huang H (2019) Incremental learning using conditional adversarial networks. In: the IEEE International Conference on Computer Vision, pp 6619--6628

\bibitem[{Xu et~al(2020)Xu, Li, Zhuang, Liu, Cao, Liang, and Tan}]{xu2020generative}
Xu S, Li H, Zhuang B, Liu J, Cao J, Liang C, Tan M (2020) Generative low-bitwidth data free quantization. In: the European Conference on Computer Vision, pp 1--17

\bibitem[{Yin et~al(2020)Yin, Molchanov, Alvarez, Li, Mallya, Hoiem, Jha, and Kautz}]{yin2020dreaming}
Yin H, Molchanov P, Alvarez JM, Li Z, Mallya A, Hoiem D, Jha NK, Kautz J (2020) Dreaming to distill: Data-free knowledge transfer via deepinversion. In: the IEEE Conference on Computer Vision and Pattern Recognition, pp 8715--8724

\bibitem[{Zagoruyko and Komodakis(2016)}]{zagoruyko2016wide}
Zagoruyko S, Komodakis N (2016) Wide residual networks. In: the British Machine Vision Conference, pp 87.1--87.12

\bibitem[{Zhang et~al(2018)Zhang, Qiao, Xie, Shen, Wang, and Yuille}]{zhang2018single}
Zhang Z, Qiao S, Xie C, Shen W, Wang B, Yuille AL (2018) Single-shot object detection with enriched semantics. In: the IEEE Conference on Computer Vision and Pattern Recognition, pp 5813--5821

\bibitem[{Zhang et~al(2022)Zhang, Luo, Wu, Chen, Wang, and Song}]{zhang2022individual}
Zhang Z, Luo C, Wu H, Chen Y, Wang N, Song C (2022) From individual to whole: Reducing intra-class variance by feature aggregation. International Journal of Computer Vision 130(3):800--819

\bibitem[{Zhong et~al(2022)Zhong, Lin, Nan, Liu, Zhang, Tian, and Ji}]{zhong2021intraq}
Zhong Y, Lin M, Nan G, Liu J, Zhang B, Tian Y, Ji R (2022) Intraq: Learning synthetic images with intra-class heterogeneity for zero-shot network quantization. In: the IEEE Conference on Computer Vision and Pattern Recognition, pp 12,339--12,348

\bibitem[{Zhu et~al(2021)Zhu, Zhang, Wang, Yin, and Liu}]{zhu2021prototype}
Zhu F, Zhang XY, Wang C, Yin F, Liu CL (2021) Prototype augmentation and self-supervision for incremental learning. In: the IEEE Conference on Computer Vision and Pattern Recognition, pp 5871--5880

\bibitem[{Zhu et~al(2022)Zhu, Zhai, Cao, Luo, and Zha}]{zhu2022self}
Zhu K, Zhai W, Cao Y, Luo J, Zha ZJ (2022) Self-sustaining representation expansion for non-exemplar class-incremental learning. In: the IEEE Conference on Computer Vision and Pattern Recognition, pp 9296--9305

\end{thebibliography}
\end{document}